\documentclass{article}

\usepackage{microtype}
\usepackage{tcolorbox}
\usepackage{float}
\usepackage[ruled,vlined]{algorithm2e} % Required for algorithms

\usepackage{graphicx}
\usepackage{subfigure}
\usepackage{multirow}
\usepackage{enumitem}
\usepackage{booktabs} % for professional tables
\usepackage{placeins}  % Add this in the preamble
\newcommand{\ours}{CoSTA$^*$}
\tcbset{
    colback=gray!10, % Light gray background
    colframe=black, % Black border
    boxrule=0.8pt, % Border thickness
    arc=3pt, % Rounded corners
}

\usepackage[bookmarks, colorlinks=true, citecolor=violet,linkcolor=brown,urlcolor=blue]{hyperref}

\usepackage[accepted]{icml2025}

\usepackage{amsmath}
\usepackage{amssymb}
\usepackage{mathtools}
\usepackage{amsthm}

\usepackage[capitalize,noabbrev]{cleveref}

\theoremstyle{plain}

\theoremstyle{definition}

\theoremstyle{remark}

\usepackage[textsize=tiny]{todonotes}

\icmltitlerunning{CoSTA*: Cost-Sensitive Toolpath Agent for Multi-turn Image Editing}

\begin{document}

\twocolumn[
\icmltitle{\ours: Cost-Sensitive Toolpath Agent for Multi-turn Image Editing}



\icmlsetsymbol{equal}{*}

\begin{icmlauthorlist}
\icmlauthor{Advait Gupta}{}
\icmlauthor{NandaKiran Velaga}{}
\icmlauthor{Dang Nguyen}{}
\icmlauthor{Tianyi Zhou}{}
\end{icmlauthorlist}

\vspace{-0.2em}
\begin{center}
% \small
% \textsuperscript{1}Johns Hopkins University;
% \textsuperscript{1}
University of Maryland, College Park\\
\texttt{\{advait25,nvelaga,dangmn,tianyi\}@umd.edu}
\end{center}

\vspace{-1em}
\begin{center}
% \texttt{zli300@jh.edu, \{litzy619,tianyi\}@umd.edu}\\
\textcolor{cyan}{Project: \url{https://github.com/tianyi-lab/CoSTAR}}
\end{center}

\icmlkeywords{Multi-turn image editing, toolpath, multimodal agent, cost-quality trade-off}

\vskip 0.3in
]




\begin{abstract}
Text-to-image models like stable diffusion and DALLE-3 still struggle with multi-turn image editing. We decompose such a task as an agentic workflow (path) of tool use that addresses a sequence of subtasks by AI tools of varying costs. 
Conventional search algorithms require expensive exploration to find tool paths. 
While large language models (LLMs) possess prior knowledge of subtask planning, they may lack accurate estimations of capabilities and costs of tools to determine which to apply in each subtask. 
\textit{Can we combine the strengths of both LLMs and graph search to find cost-efficient tool paths?}
We propose a three-stage approach ``\ours{}'' that leverages LLMs to create a subtask tree, which helps prune a graph of AI tools for the given task, and then conducts A$^*$ search on the small subgraph to find a tool path. 
To better balance the total cost and quality, \ours{} combines both metrics of each tool on every subtask to guide the A$^*$ search. Each subtask's output is then evaluated by a vision-language model (VLM), where a failure will trigger an update of the tool's cost and quality on the subtask. 
Hence, the A$^*$ search can recover from failures quickly to explore other paths. 
Moreover, \ours{} can automatically switch between modalities across subtasks for a better cost-quality trade-off. 
We build a novel benchmark of challenging multi-turn image editing, on which \ours{} outperforms state-of-the-art image-editing models or agents in terms of both cost and quality, and performs versatile trade-offs upon user preference. 
% Our dataset and a hosted demo can be found \href{https://storage.googleapis.com/costa-frontend/index.html}{here}.
\end{abstract}

\section{Introduction}
\label{sec:intro}
% \FloatBarrier
\begin{figure}[ht]
    \centering
    \includegraphics[width=\linewidth]{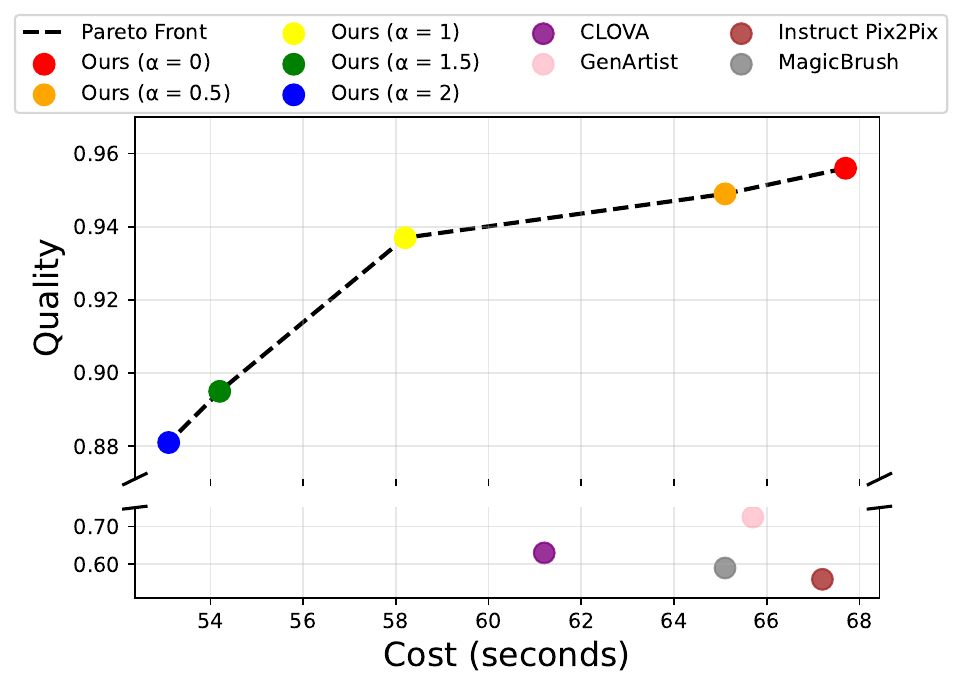}
    \vspace{-2em}
    \caption{\ours{} with different cost-quality trade-off coefficients $\alpha$ vs. four recent image-editing models/agents. \ours{} achieves Pareto optimality and dominates baselines on both metrics.}
    \label{fig:pareto}
    \vspace{-1em}
\end{figure}
% \paragraph{Background}
Text-to-Image models such as stable diffusion, FLUX, and DALLE has been widely studied to replace humans on image-editing tasks, which are time-consuming due to various repetitive operations and trial-and-errors. While these models have exhibited remarkable potential for generating diverse images and simple object editing, they usually struggle to follow composite instructions that require multi-turn editing, in which a sequence of delicate adjustments are requested to manipulate (e.g., remove, replace, add) several details (e.g., object attributes or texts) while keeping other parts intact. For example, given an image, it is usually challenging for them to ``recolor the chalkboard to red while redacting the text on it and write “A CLASSROOM” on the top. Also, detect if any children are in the image.''
\begin{figure*}[t]
    \centering
    \includegraphics[width=1\textwidth]{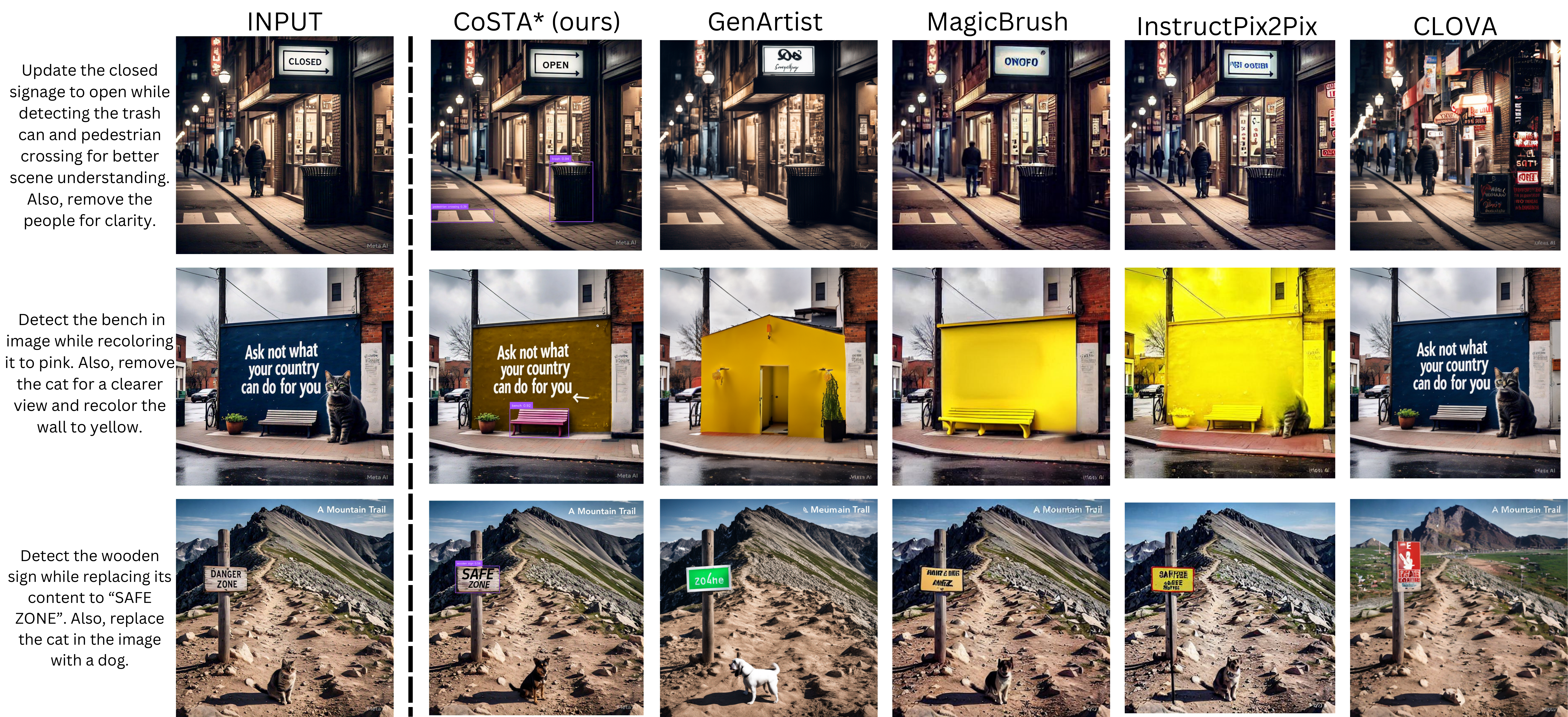}
        \vspace{-1em}
    \caption{Comparison of \ours{} with State-of-the-Art image editing models/agents, which include GenArtist \cite{wang2024genartistmultimodalllmagent}, MagicBrush \cite{zhang2024magicbrushmanuallyannotateddataset}, InstructPix2Pix \cite{brooks2023instructpix2pixlearningfollowimage}, and CLOVA \cite{DBLP:conf/cvpr/Gao0ZMHZL24}. The input images and prompts are shown on the left of the figure. The outputs generated by each method illustrate differences in accuracy, visual coherence, and the ability to multimodal tasks. Figure~\ref{fig:qual_example} shows examples of step-by-step editing using \ours with intermediate subtask outputs presented.}
    \label{fig:qual_comp}
    \vspace{-1em}
\end{figure*}
% Image editing and text-in-image editing are needed in a lot of practical applications. sssWhile recent text-to-image models such as stable diffusion and DALLE has shown great potential on these tasks, they still struggle on composite editing tasks such as
% \nanda{felt like a cut in flow of reading}
% \paragraph{Challenges}

Although a large language model (LLM) can decompose the above multi-turn composite task into easier subtasks, and each subtask can be potentially learned by existing techniques such as ControlNet, the required training data and computational costs are usually expensive. Hence, a training-free agent that automatically selects tools to address the subtasks is usually more appealing. However, finding an efficient and successful path of tool use (i.e., toolpath) is nontrivial: while some subtasks are exceptionally challenging and may require multi-round trial-and-errors with advanced and costly AI models, various subtasks could be handled by much simpler, lower-cost tools. Moreover, users with limited budgets usually prefer to control and optimize the trade-off between quality and cost. However, most existing image-editing agents are not cost-sensitive so the search cost of their toolpaths can be highly expensive. 

\begin{figure*}[ht]  % Place at the top of the page
    \centering
    \includegraphics[width=0.88\textwidth]{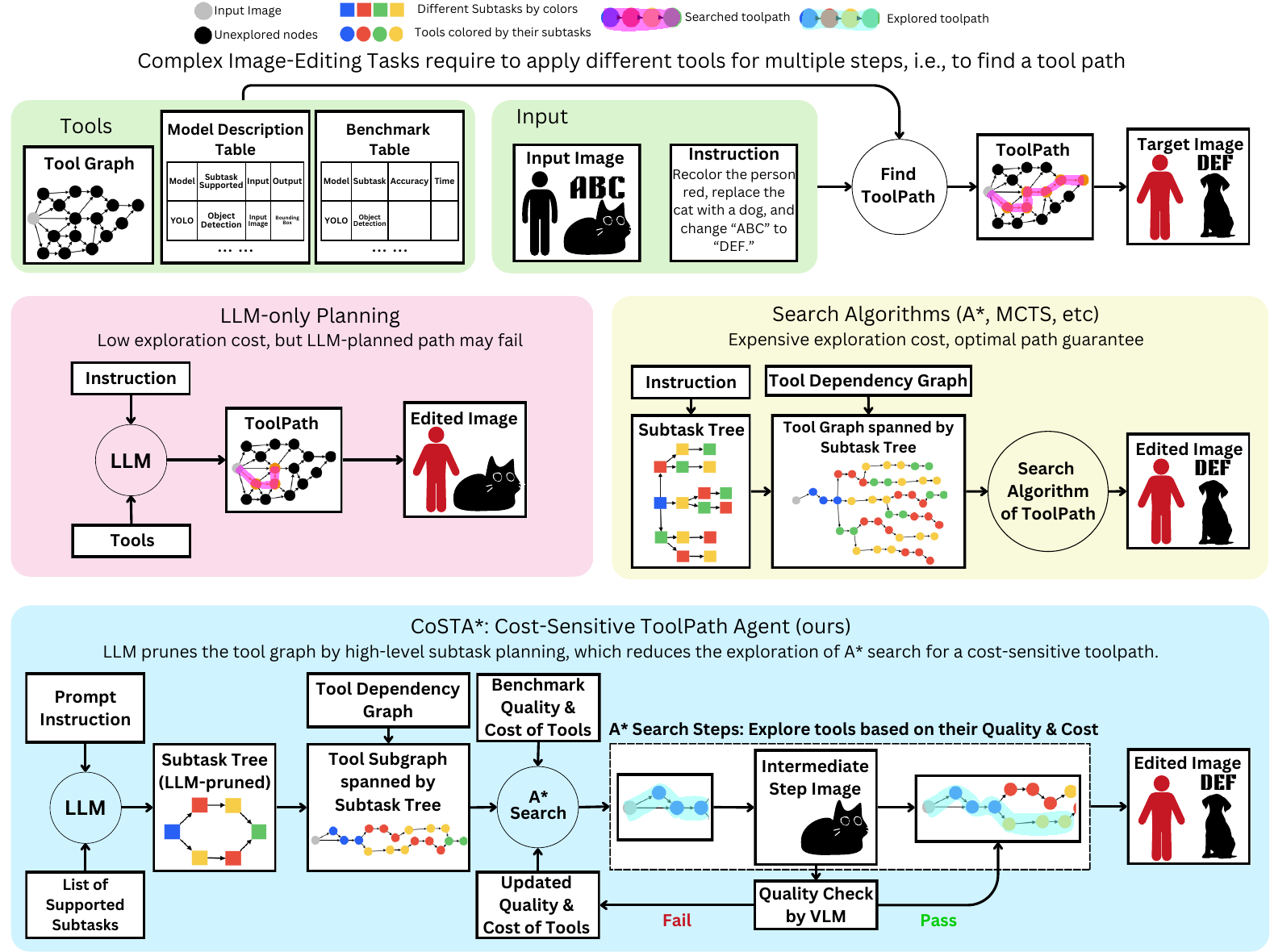}
    \caption{Comparison of \ours{} with other planning agents. LLM-only planning is efficient but prone to failure and heuristics. Search algorithms like A$^*$ guarantee optimal paths but are computationally expensive. \ours{} balances cost and quality by first pruning the subtask tree using an LLM, which reduces the graph of tools we conduct fine-grained A$^*$ search on.  
    % enabling a more efficient and reliable tool selection process.
    }
    \label{fig:novelty}
\end{figure*}

Despite the strong heuristic of LLMs on tool selection for each subtask, as shown in Figure~\ref{fig:novelty}, they also suffer from hallucinations and may generate sub-optimal paths due to the lack of precise knowledge for each tool and the long horizon of multi-turn editing. On the other hand, classical search algorithms such as A$^*$ and MCTS can precisely find the optimal tool path after sufficient exploration, if accurate estimates of per-step value/cost and high-quality heuristics are available. However, they are not scalable to explore tool paths on a large-scale graph of many computationally heavy models as tools, e.g., diffusion models. This motivates the question: \textit{Can we combine the strengths of both methods in a complementary manner?}
% \nanda{Existing methods cost mentioning in the background or here}

\begin{figure*}[t]
    \centering
    \includegraphics[width=1\textwidth]{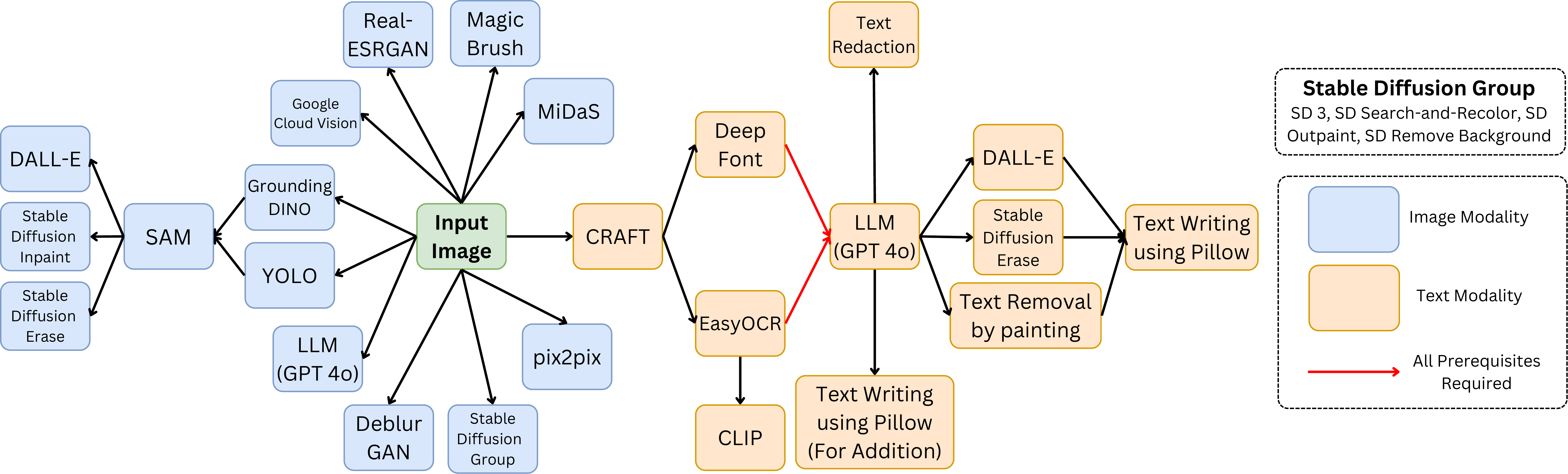}
    \vspace{-1em}
    \caption{\textbf{Tool Dependency Graph (TDG)}. A directed graph where nodes represent tools and edges indicate dependencies. An edge \( (v_1, v_2) \) means \( v_1 \)'s output is a legal input of \( v_2 \). It enables toolpath search for multi-turn image-editing tasks with composite instructions.\looseness-1}
    \label{fig:tdg}
    \vspace{-1.5em}
\end{figure*}

% \paragraph{Motivations}
In this paper, we develop a novel agentic mechanism ``\textbf{Cost-Sensitive Toolpath Agent (\ours{})}'' that combines both LLMs and A$^*$ search's strengths while overcoming each other's weaknesses to find a cost-sensitive path of tool use for a given task. 
As illustrated in Figure~\ref{fig:novelty}, we propose a hierarchical planning strategy where an LLM focuses on subtask planning (each subtask is a subsequence of tool uses), which decomposes the given task into a subtask tree on which every root-to-leaf path is a feasible high-level plan for the task. 
This is motivated by the observations that LLMs are more powerful on subtask-level commonsense reasoning but may lack accurate knowledge to decide which specific tools to use per subtask. 
Then, a low-level A$^*$ search is applied to the subgraph spanned by the subtask tree on a tool dependency graph (TDG, with an example in Figure~\ref{fig:tdg}). It aims to find a toolpath fulfilling the user-defined quality-cost trade-off. 
The subtask tree effectively reduces the graph of tools on which the A$^*$ search is conducted, saving a significant amount of searching cost. \looseness-1
% Therefore, we use an LLM to decompose a composite task into a subtask tree, where each root-to-leaf path provides a feasible plan of subtasks to accomplish the original task. We then conduct A$^*$ search on a much smaller subgraph of tools . 

In \ours{}, we exploit available prior knowledge and benchmark evaluation results of tools, which are underexplored in previous LLM agents, to improve both the planning and search accuracy. We mainly leverage two types of prior information: (1) the input, output, and subtasks of each tool/model; and (2) the benchmark performance and cost of each tool or model reported in the existing literature. 
Specifically, a sparse tool dependency graph (TDG) is built based on (1), where two tools are connected if the first's output is a legal input to the second in certain subtask(s). Moreover, the information in (2) defines the heuristics $h(x)$ in A$^*$ search, which combines both the cost and quality with a trade-off coefficient $\alpha$. We further propose an actual execution cost $g(x)$ combining the actual cost and quality in completed subtasks, and update it during exploration. By adjusting $\alpha$, the cost-sensitive A$^*$ search aims to find a toolpath aligning with user preference of quality-cost trade-off. 
% As Our low-level toolpath search is conducted on a subgraph of the TDG, which covers all the subtasks in the LLM-generated planning tree. Moreover,                                          

% Thirdly, we study a cost-sensitive agentic mechanism, as many simpler editing tasks can be handled by more efficient tool paths that do not require expensive image generation. 

% \paragraph{Our contributions}

To examine the performance of \ours{}, we curate a novel benchmark for multi-turn image editing with challenging, composite tasks. We compare \ours{} with state-of-the-art image-editing models or agents. As shown in Figure~\ref{fig:pareto}, \ours achieves advantages over others on both the cost and quality, pushing the Pareto frontier of their trade-offs. In Figure~\ref{fig:qual_example}, in several challenging multi-turn image-editing tasks, only \ours{} accomplishes the goals. 
Our contributions can be summarized as below:
\begin{itemize}[leftmargin=1em, itemsep=0.1em]
\vspace{-0.5em}
\item We propose a novel hierarchical planning agent \ours{} that combines the strengths of LLMs and graph search to find toolpaths for composite multi-turn image editing.
\item \ours{} addresses the quality-cost trade-off problem by a controllable cost-sensitive A$^*$ search, and achieves the Pareto optimality over existing agents.
\item We exploit prior knowledge of tools to improve the toolpath finding. 
\item We propose a new challenging benchmark for multi-turn image editing covering tasks of different complexities. 
%     \item We developed a novel agentic mechanism \ours{} to combine the strengths of LLMs and classical graph search algorithms with hierarchical planning, for efficient search of high-quality paths for tool use (ToolPath). It avoids the expensive cost of graph search on the complete tool graph by only conducting the precise A$^*$ search on a subtask tree pruned by an LLM. 
%     \item \ours{} can recover from failures or a suboptimal path. Its planning also leverages existing knowledge of tools such as their benchmark performance and computational cost on compatible subtasks. 
%     \item \ours{} is cost-sensitive and can adjust the trade-off between quality and cost to achieve different ToolPath solutions on the Pareto front. 
%     \item \ours{} can handle more diverse and composite tasks not fully covered by existing image-editing agents. It can explore the best modality to perform each subtask and switch between modalities based on visual feedback if necessary. 
%     \item Figure \ref{fig:novelty} highlights our novelties and compares our approach with other standard approaches currently being used for such editing tasks.
\end{itemize}

\section{Related Work}
\label{sec:related}

\textbf{Image Editing via Generative AI }
Image editing has seen significant advancements with the rise of diffusion models \cite{DBLP:conf/nips/DhariwalN21, DBLP:conf/nips/HoJA20}, enabling highly realistic and diverse image generation and modification. Modern approaches focus on text-to-image frameworks that transform descriptive text prompts into images, achieving notable quality \cite{chen2023pixartalphafasttrainingdiffusion, rombach2022highresolutionimagesynthesislatent, saharia2022photorealistictexttoimagediffusionmodels} but often facing challenges with precise control over outputs. To mitigate this controllability issue, methods like ControlNet \cite{zhang2023addingconditionalcontroltexttoimage} and sketch-based conditioning \cite{voynov2022sketchguidedtexttoimagediffusionmodels} refine user-driven edits, while layout-to-image systems synthesize compositions from spatial object arrangements \cite{chen2023trainingfreelayoutcontrolcrossattention, li2023gligenopensetgroundedtexttoimage, lian2024llmgroundeddiffusionenhancingprompt, xie2023boxdifftexttoimagesynthesistrainingfree}. Beyond text-driven editing, research efforts have also focused on personalized generation and domain-specific fine-tuning for tasks such as custom content creation or rendering text within images. However, current models still struggle with handling complex prompts, underscoring the need for unified, flexible solutions \cite{brooks2023instructpix2pixlearningfollowimage, chen2024anydoorzeroshotobjectlevelimage, parmar2023zeroshotimagetoimagetranslation, yang2022paintexampleexemplarbasedimage}.

\textbf{Large Multimodal Agents for Image Editing }
Recent advancements in multimodal large language models (MLLMs) have significantly enhanced complex image editing capabilities \cite{wang2024genartistmultimodalllmagent, huang2024dialoggenmultimodalinteractivedialogue, zhang2024itercompiterativecompositionawarefeedback, huang2023smarteditexploringcomplexinstructionbased, zhang2024sgeditbridgingllmtext2image, yang2024idea2imgiterativeselfrefinementgpt4vision, wang2024divideconquerlanguagemodels}. GenArtist \cite{wang2024genartistmultimodalllmagent} introduces a unified system where an MLLM agent coordinates various models to decompose intricate tasks into manageable sub-problems, enabling systematic planning and self-correction.  DialogGen \cite{huang2024dialoggenmultimodalinteractivedialogue} aligns MLLMs with text-to-image (T2I) models, facilitating multi-turn dialogues that allow users to iteratively refine images through natural language instructions. IterComp \cite{zhang2024itercompiterativecompositionawarefeedback} aggregates preferences from multiple models and employs iterative feedback learning to enhance compositional generation, particularly in attribute binding and spatial relationships. SmartEdit \cite{huang2023smarteditexploringcomplexinstructionbased} leverages MLLMs for complex instruction-based editing, utilizing a bidirectional interaction module to improve understanding and reasoning. These approaches build upon foundational works like BLIP-2 \cite{li2023blip2bootstrappinglanguageimagepretraining}, which integrates vision and language models for image understanding, and InstructPix2Pix \cite{brooks2023instructpix2pixlearningfollowimage}, which focuses on text-guided image editing.

\vspace{-1em}

\section{Foundations of \ours{}}
\vspace{-0.7em}

% This section outlines the foundational resources, prior knowledge, and components utilized in \ours{} to support efficient and cost-sensitive multimodal editing tasks.
We present the underlying models, supporting data structures, and prior knowledge that \ours{} relies on before explaining the design of the \ours{} algorithm. Specifically, we describe the Model Description Table, the Tool Dependency Graph, and the Benchmark Table.

\vspace{-0.5em}
\subsection{Model Description Table}
\vspace{-1.5em}
\begin{table}[h]
    \centering
    \caption{Model Description Table (excerpt)}
    \label{tab:mdt}
    \resizebox{\columnwidth}{!}{%
        \begin{tabular}{l l l l}
            \toprule
            Model & Supported Subtasks & Inputs & Outputs \\
            \midrule
            YOLO~\cite{wang2022yolov7trainablebagoffreebiessets} & Object Detection & Input Image & Bounding Boxes \\
            
            SAM~\cite{DBLP:conf/iccv/KirillovMRMRGXW23} & Segmentation & Bounding Boxes & Segmentation Masks \\
            
            DALL-E~\cite{DBLP:journals/corr/abs-2102-12092} & Object Replacement & Segmentation Mask & Edited Image \\
            
            Stable Diffusion Inpaint & Object Removal, Replacement, & Segmentation Mask & Edited Image \\

            ~\cite{DBLP:conf/cvpr/RombachBLEO22} & Recoloration & & \\
            
            EasyOCR & Text Extraction & Text Bounding Box & Extracted Text \\

            ~\cite{rakpong_kittinaradorn_2022_6850706} & & & \\
            \bottomrule
        \end{tabular}
    }
    \vspace{-0.5em}
\end{table}

% Our corpus contains \(2\) specialized models in the \ours{} pipeline for image and text-in-image editing. Models are accessed via APIs when available; otherwise, we use downloaded weights or trained models. The Model Description Table (MDT) lists each model \( v_i \in V_{\text{td}} \) with its supported subtasks \( \mathcal{T}_{v_i} \subseteq \mathcal{S} \).

% Unlike generic frameworks, our approach leverages specialized models for text-in-image tasks, ensuring high accuracy and efficiency. Table~\ref{tab:mdt} provides an excerpt of the MDT, illustrating key model capabilities.

We first construct a Model Description Table (MDT) that lists all specialized models (e.g., SAM, YOLO) and the corresponding tasks they support (e.g., image segmentation, object detection). In this paper, we consider 24 models that collectively support 24 tasks, covering both image and text modalities. The supported tasks can be broadly categorized into \emph{image editing} tasks (e.g., object removal, object recolorization) and \emph{text-in-image editing} tasks (e.g., text removal, text replacement). Our system allows for easy extension by adding new models and their corresponding tasks to this table. The MDT also includes columns specifying the input dependencies and outputs of each model. An excerpt of the MDT is shown in Table \ref{tab:mdt} to illustrate its structure, and full MDT is available in Appendix (Table \ref{tab:full_mdt}).

% \subsubsection{Tool Dependency Graph}
% \label{sec:tool_dep_graph}
% We consider each tool in our tool library as a different model specialized in handling a specific type of task. For example, YOLO \cite{DBLP:conf/cvpr/RedmonDGF16} is a tool for object detection, SAM \cite{DBLP:conf/iccv/KirillovMRMRGXW23} is for segmentation, and DALL-E \cite{DBLP:journals/corr/abs-2102-12092} is for image editing.

% We define the \emph{tool dependency graph (TDG)} as a directed graph $G_\text{td} = (V_\text{td}, E_\text{td})$ that represents the dependency relationship between different tools in the library, where $V_\text{td}$ is the set of all tools, $E_{\text{td}} \subseteq V_{\text{td}} \times V_{\text{td}}$ is the set of directed edges such that $(v_1, v_2) \in E_\text{td}$ if and only if the tool $v_2$ depends on the output of tool $v_1$. The tool dependency graph for our study is shown in Figure \ref{fig:tdg}.
\vspace{-1em}
\subsection{Tool Dependency Graph}
\label{sec:tool_dep_graph}
\vspace{-0.5em}
Each tool in our library is a specialized model for a specific subtask, where some tools require the outputs of other tools as inputs. To capture these dependencies, we construct a Tool Dependency Graph (TDG). Formally, we define the TDG as a directed graph $G_{\text{td}} = (V_{\text{td}}, E_{\text{td}})$, where $V_{\text{td}}$ is the set of tools, and $E_{\text{td}} \subseteq V_{\text{td}} \times V_{\text{td}}$ contains edges $(v_1, v_2)$ if tool $v_2$ depends on the output of $v_1$. Figure~\ref{fig:tdg} presents the full TDG, illustrating the dependencies between tools. This TDG can be automatically generated based on the input-output specifications of each tool mentioned in the MDT, reducing the need for extensive human effort (see Appendix~\ref{sec:tdg_auto} for a detailed explanation).
\vspace{-1em}

\subsection{Benchmark Table for Heuristic Scores}
\vspace{-0.5em}

% The Benchmark Table (BT) defines predefined time and quality values for each tool-task pair \( \text{BT}(v_i, s_j) \), where \( v_i \in V_{\text{td}} \) is a tool (model) and \( s_j \in S \) is a subtask. It serves as the baseline for the heuristic \( h(x) \), enabling A$^*$ to estimate tool cost-effectiveness. BT values are sourced from published studies and benchmarks (e.g., mAP for YOLO \cite{DBLP:conf/cvpr/RedmonDGF16}, F1 for CRAFT \cite{DBLP:conf/cvpr/BaekLHYL19}), with execution times standardized. For tools lacking benchmarks, evaluations on 100–150 samples are conducted and their details noted. Each tool-task pair has distinct time \( C(v_i, s_j) \) and quality \( Q(v_i, s_j) \) values.  Quality values are normalized per subtask (0-1 scale) for comparability. \looseness-1

\begin{figure*}[t]
    \centering
    \includegraphics[width=1\textwidth]{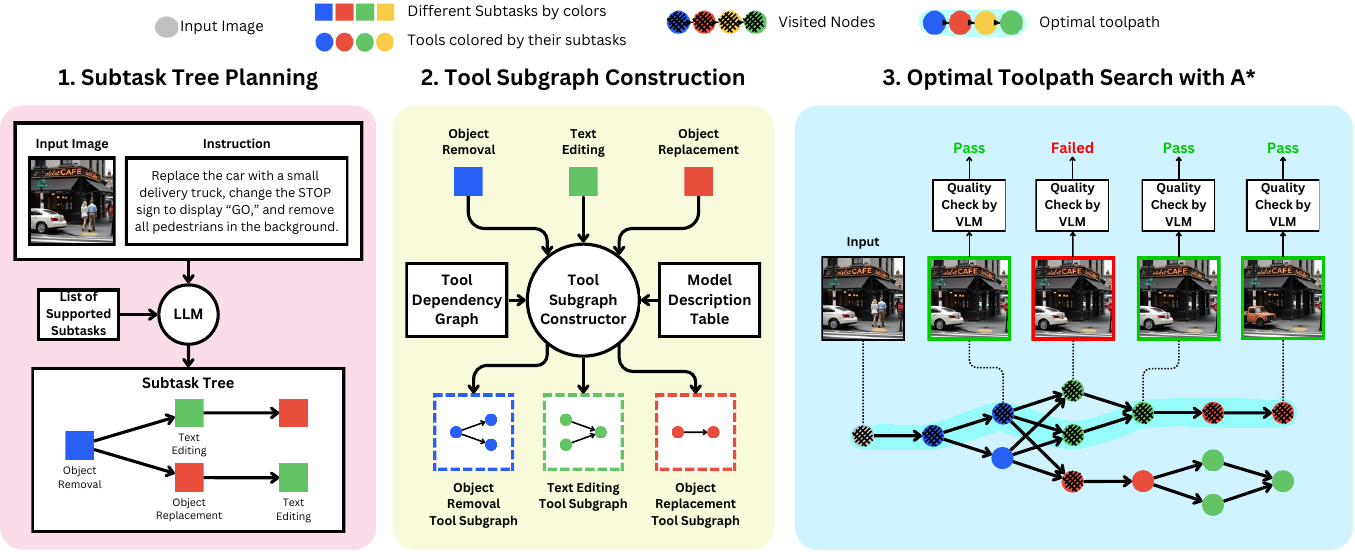}
    \vspace{-1em}

    \caption{\textbf{Three stages in \ours{}}: (1) an LLM generates a subtask tree based on the input and task dependencies; (2) the subtask tree spans a tool subgraph that maintains tool dependencies; and (3) A$^*$ search finds the best toolpath balancing efficiency and quality.\looseness-1}
    \label{fig:main}
    \vspace{-1.em}
\end{figure*}

At its core, \ours{} employs A$^*$ search over a network of interdependent tools to find the optimal cost-sensitive path. This process relies on a heuristic function $h(x)$ for each tool $x$. We initialize these heuristic values using prior knowledge of execution time and quality scores obtained from existing benchmarks or published studies (e.g., mAP score for YOLO \cite{wang2022yolov7trainablebagoffreebiessets} and F1 score for CRAFT \cite{DBLP:conf/cvpr/BaekLHYL19}). Since not all tools have sufficient benchmark data, we evaluate them over \textbf{137 instances of the specific subtask}, applied across \textbf{121 images from the dataset} to handle missing values. For each tool-task pair $(v_i, s_j)$, we define an execution time $C(v_i, s_j)$ and a quality score $Q(v_i, s_j)$. To ensure comparability, quality values are normalized per subtask to a $[0,1]$ scale. The complete Benchmark Table (BT) is shown in Table \ref{tab:bt}.
\vspace{-1em}

% \begin{figure*}[t]
%     \centering
%     \includegraphics[width=1\textwidth]{sec/main.pdf}
%     \vspace{-1em}

%     \caption{\textbf{Three stages in \ours{}}: (1) an LLM generates a subtask tree based on the input and task dependencies; (2) the subtask tree spans a tool subgraph that maintains tool dependencies; and (3) A$^*$ search finds the best toolpath balancing efficiency and quality.\looseness-1}
%     \label{fig:main}
%     \vspace{-1.em}
% \end{figure*}

\section{\ours{}: Cost-Sensitive Toolpath Agent} \label{sec:methodology}
\vspace{-0.5em}

% \begin{figure*}[t]
%     \centering
%     \includegraphics[width=1\textwidth]{sec/main.pdf}
%     \vspace{-1em}

%     \caption{Overview of our method. Our approach consists of three stages: (1) an LLM generates a subtask tree based on the input and task dependencies, (2) the subtask tree is mapped to a Tool Subgraph that maintains model dependencies, and (3) A* search finds the best execution path by balancing efficiency and quality.}
%     \label{fig:main}
%     \vspace{-1em}

% \end{figure*}

This section details our approach for constructing and optimizing a Tool Subgraph (TS) to efficiently execute multimodal editing tasks. The methodology consists of three key stages: (1) generating a subtask tree, (2) constructing the TS, and (3) applying A$^*$ search to determine the optimal execution path.

% \begin{figure*}[t]
%     \centering
%     \includegraphics[width=1\textwidth]{sec/main.pdf}
%     \vspace{-1em}

%     \caption{Overview of our method. Our approach consists of three stages: (1) an LLM generates a subtask tree based on the input and task dependencies, (2) the subtask tree is mapped to a Tool Subgraph that maintains model dependencies, and (3) A* search finds the best execution path by balancing efficiency and quality.}
%     \label{fig:main}
%     \vspace{-1.5em}

% \end{figure*}

First, as shown in Figure~\ref{fig:main}, an LLM infers subtasks and dependencies from the input image, prompt, and the set of supported subtasks \( \mathcal{S} \), generating a subtask tree \( G_\text{ss} \). Then, this tree is transformed into the Tool Subgraph \( G_\text{ts} \), where each subtask is mapped to a model subgraph within the TDG. This ensures that model dependencies are maintained while incorporating task sequences and execution constraints. Finally, A$^*$ search explores \( G_\text{ts} \) to identify an optimal execution path by balancing computational cost and output quality. It prioritizes paths based on a cost function $f(x) = g(x) + h(x)$ where \( g(x) \) represents real-time execution costs, and \( h(x) \) is the precomputed heuristic. A tunable parameter \( \alpha \) controls the tradeoff between efficiency and quality, allowing for adaptive optimization.

\vspace{-0.5em}

% \begin{figure*}[t]
%     \centering
%     \includegraphics[width=1\textwidth]{sec/main.pdf}
%     \vspace{-1em}

%     \caption{Overview of our method. Our approach consists of three stages: (1) an LLM generates a subtask tree based on the input and task dependencies, (2) the subtask tree is mapped to a Tool Subgraph that maintains model dependencies, and (3) A* search finds the best execution path by balancing efficiency and quality.}
%     \label{fig:main}
%     \vspace{-1.5em}

% \end{figure*}
\vspace{-0.5em}
\subsection{Task Decomposition \& Subtask Tree Planning} \label{sec:task_decomp_subtask_plan}
\vspace{-0.5em}

Given an input image \(x\) and prompt \(u\), we employ an LLM \(\pi(\cdot | f_{\text{plan}}(x, u, \mathcal{S}))\) to generate a subtask tree \( G_\text{ss} = (V_\text{ss}, E_\text{ss}) \), where each node \( v_i \) represents a subtask \( s_i \), and each edge \((v_i, v_j)\) denotes a dependency. Here, \( f_{\text{plan}} \) is a prompt template containing the input image, task description \( u \), and supported subtasks \( \mathcal{S} \). The full prompt is detailed in Appendix \ref{prompt1}. The LLM infers task relationships, forming a directed acyclic graph where each root-to-leaf path represents a valid solution.

The subtask tree encodes various solution approaches, accommodating different subtask orders and workflows. Path selection determines an optimized workflow based on efficiency or quality. Part 1 of Figure~\ref{fig:main} (Subtask Tree Planning) illustrates an example where the LLM constructs a subtask tree from an input image and prompt.

\subsection{Tool Subgraph Construction}

The TS, denoted as \( G_{ts} = (V_{ts}, E_{ts}) \), represents the structured execution paths for fulfilling subtasks in the \emph{Subtask Tree} (ST) \( G_{\text{ss}} \). It is constructed by mapping each subtask node to a corresponding model subgraph from the TDG \( G_{td} \).\looseness-1

The \emph{node set} \( V_{ts} \) consists of all models required for execution, ensuring that every subtask \( s_i \in S \) is associated with a valid model:
\begin{equation}
V_{ts} = \bigcup_{s_i \in S} M(s_i),
\end{equation}
where \( M(s_i) \) denotes the set of models that can perform subtask \( s_i \), as listed in the MDT.

\begin{figure*}[t]
    \centering
    \includegraphics[width=1\linewidth]{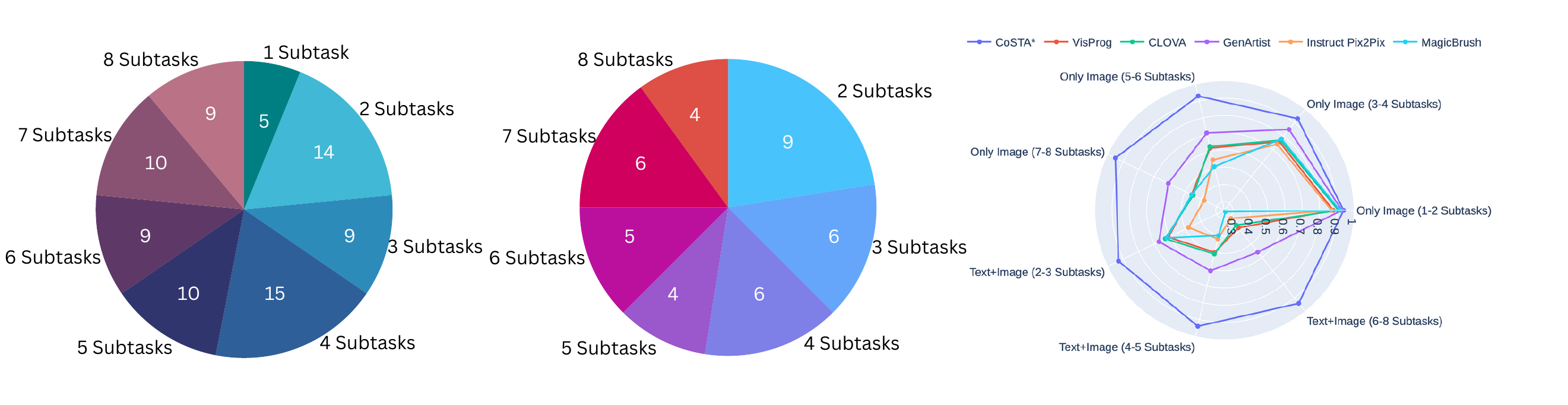}
    \vspace{-2.5em}
    \caption{Distribution of image-only (left) and text+image tasks (middle) in \textbf{our proposed benchmark, and quality comparison} of different methods on the benchmark (right). \ours{} excels in complex multimodal tasks and outperforms all the baselines. 
    % with the radar plot highlighting its scalability.
    }
    \label{fig:dataset}
    \vspace{-0.5em}
\end{figure*}

The \emph{edge set} \( E_{ts} \) represents dependencies between models, ensuring that each model receives the necessary inputs from its predecessors before execution. These dependencies are derived from \( G_{td} \) by backtracking to identify required intermediate outputs:
\begin{equation}
E_{ts} = \bigcup_{s_i \in S} E_{ti},
\end{equation}
where \( E_{ti} \) contains directed edges between models in \( M(s_i) \) based on their execution dependencies.

The final tool subgraph \( G_{ts} \) encapsulates all feasible execution paths while preserving dependencies and logical consistency. Figure~\ref{fig:main} (Tool Subgraph Construction) illustrates this transformation.

% \begin{figure*}[t]
%     \centering
%     \includegraphics[width=1\linewidth]{sec/combine.pdf}
%     \vspace{-2.5em}
%     \caption{Dataset distribution for image-only (left) and text+image tasks (center) and quality comparison of different methods (right). \ours{} excels in complex multimodal tasks, with the radar plot highlighting its scalability.}
%     \label{fig:dataset}
%     \vspace{-0.5em}
% \end{figure*}
% \vspace{-1em}
\subsection{Path Optimization with A$^*$ Search}
The A$^*$ algorithm finds the optimal execution path by minimizing the cost function: $f(x) = g(x) + h(x)$ where \( g(x) \) is the \textbf{actual execution cost}, dynamically updated during execution, and \( h(x) \) is the \textbf{heuristic estimate}, precomputed from benchmark values. Nodes are explored in increasing order of \( f(x) \), ensuring an efficient tradeoff between execution time and quality.
\vspace{-1em}

\subsection{Heuristic Cost $h(x)$}
The heuristic cost \( h(x) \) estimates the best-case execution cost from node \( x \) to a leaf node (excluding the cost of \( x \) itself), factoring in both execution time and quality. Each node represents a tool-task pair \( (v_i, s_i) \), where \( v_i \) is the tool and \( s_i \) is the subtask. For example, \( y = (\text{YOLO}, \text{Object Detection}) \) ensures that \( y \) is inherently multivariate. The heuristic is defined as:
\begin{equation*}
\begin{aligned}
h(x) = \min_{y \in \text{Neighbors}(x)} & \left( (h_C(y) + C(y))^\alpha \right. \\
& \times \left. (2- Q(y) \times h_Q(y))^{(2-\alpha)} \right)
\end{aligned}
\end{equation*}
where \( h_C(y) \) represents the cost component of \( h(y) \) (initialized as 0 for leaf nodes), while \( h_Q(y) \) denotes the quality component (initialized as 1 for leaf nodes). \( C(y) \) and \( Q(y) \) correspond to the benchmark execution time and quality of tool \( y \), respectively, and \( \alpha \) controls the tradeoff between cost and quality. This heuristic propagates recursively, ensuring each node maintains the best possible estimate to a leaf node.
\vspace{-2em}

\subsection{Actual Execution Cost \( g(x) \)}
\vspace{-0.5em}

The actual execution cost \( g(x) \) is computed in real-time as execution progresses:
\vspace{-1em}
\begin{equation}
    g(x) = \left( \sum_{i=1}^{x} c(v_i, s_i) \right)^\alpha \times \left(2 - \prod_{i=1}^{x} q(v_i, s_i) \right)^{2 - \alpha}
\end{equation}
\vspace{0em}
where \( c(v_i, s_i) \) represents the actual execution time (in seconds) of the tool-subtask pair \( (v_i, s_i) \), and \( q(v_i, s_i) \) is the real-time validated quality score for the same pair. 

The summation includes only nodes in the currently explored path. Each node is initialized with \( g(x) = \infty \), except the start node, which is set to zero. Upon execution, \( g(x) \) is updated to the minimum observed value.

If a node \( x \) fails the manually set quality threshold, it undergoes a retry mechanism with updated hyperparameters. If successful, the new execution cost is accumulated in \( g(x) \). If a node fails all retries, \( g(x) \) remains unchanged, and the path is not added back to the queue, ensuring failed paths are deprioritized, but alternative routes exploring the same node remain possible. More information about the execution is in Appendix \ref{execution}.\looseness-1

\vspace{-1em}

\section{Experiments} \label{sec:experiments}
\vspace{-0.5em}
We evaluate \ours{} on a curated dataset, comparing it against baselines to assess its effectiveness in complex image and text-in-image editing. This section details experimental settings, results, ablation studies, and case studies showcasing \ours{}’s capabilities.
\vspace{-1em}
\subsection{Experimental Settings} \label{sec:experimental_settings}
\vspace{-0.5em}

% \subsubsection{Benchmark Dataset} \label{sec:dataset}
\textbf{Benchmark Dataset }
Our dataset consists of 121 manually curated images with prompts involving 1–8 subtasks per task, ensuring comprehensive coverage across both image and text-in-image modalities. It includes 81 tasks with image-only edits and 40 tasks requiring multimodal processing. The dataset is evenly distributed across subtask counts. Figure~\ref{fig:dataset} summarizes its distribution, with further details in Appendix~\ref{Dataset}.
\vspace{-0.5em}

\textbf{Baselines }
We compare \ours{} against agentic baselines such as VISPROG~\cite{DBLP:conf/cvpr/GuptaK23}, GenArtist~\cite{DBLP:journals/corr/abs-2407-05600}, and CLOVA~\cite{DBLP:conf/cvpr/Gao0ZMHZL24}. These methods support task orchestration but lack \ours{}’s A$^*$ path optimization, cost-quality tradeoff, and multimodal capabilities. Additionally, they handle only 5–6 subtasks, limiting flexibility, especially for text-in-image editing, compared to \ours{}’s 24 supported subtasks.

\vspace{-1em}
\subsection{Evaluation Metrics}

\begin{table*}[t]
\vspace{-1em}

    \centering
    \caption{Accuracy comparison of \ours{} with baselines across task types and categories. \ours{} excels in complex workflows with A$^*$ search and a diverse set of tools, ensuring higher accuracy.}
    \small
    \label{tab:accuracy_comparison}
        \resizebox{\textwidth}{!}{%

    \begin{tabular}{l l ccccc c}
        \toprule
        \textbf{Task Type} & \textbf{Task Category} & \textbf{\ours{}} & \textbf{VisProg} 
         & \textbf{CLOVA} & \textbf{GenArtist} & \textbf{Instruct Pix2Pix} & \textbf{MagicBrush}\\
             & & & \hspace{-1em}{\scriptsize\cite{DBLP:conf/cvpr/GuptaK23}} & \hspace{-1em}{\scriptsize \cite{DBLP:conf/cvpr/Gao0ZMHZL24}} & {\scriptsize\cite{wang2024genartistmultimodalllmagent}} & {\scriptsize\cite{brooks2023instructpix2pixlearningfollowimage}} & {\scriptsize\cite{DBLP:conf/nips/ZhangMCSS23}} \\
        \midrule
        \multirow{4}{*}{\textbf{Image-Only Tasks}}  
        & 1--2 subtasks & \textbf{0.94} & 0.88 & 0.91 & 0.93 & 0.87 & 0.92 \\
        & 3--4 subtasks & \textbf{0.93} & 0.76 & 0.77 & 0.85 & 0.74 & 0.78 \\
        & 5--6 subtasks & \textbf{0.93} & 0.62 & 0.63 & 0.71 & 0.55 & 0.51 \\
        & 7--8 subtasks & \textbf{0.95} & 0.46 & 0.45 & 0.61 & 0.38 & 0.46 \\
        \midrule
        \multirow{3}{*}{\textbf{Text+Image Tasks}}  
        & 2--3 subtasks & \textbf{0.93} & 0.61 & 0.63 & 0.67 & 0.48 & 0.62 \\
        & 4--5 subtasks & \textbf{0.94} & 0.50 & 0.51 & 0.61 & 0.42 & 0.40 \\
        & 6--8 subtasks & \textbf{0.94} & 0.38 & 0.36 & 0.56 & 0.31 & 0.26 \\
        \midrule
        \multirow{3}{*}{\textbf{Overall Accuracy}}  
        & Image Tasks & \textbf{0.94} & 0.69 & 0.70 & 0.78 & 0.64 & 0.67 \\
        & Text+Image Tasks & \textbf{0.93} & 0.49 & 0.50 & 0.61 & 0.40 & 0.43 \\
        & All Tasks & \textbf{0.94} & 0.62 & 0.63 & 0.73 & 0.56 & 0.59\\
        \bottomrule
    \end{tabular}%
    }
    
    \vspace{-1em}

\end{table*}
\vspace{-0.5em}

\paragraph{Human Evaluation} \label{accuracy}

To ensure a reliable assessment of model performance, we employ human evaluation for accuracy measurement. Each subtask \( s_i \) in task \( T \) is manually assessed and assigned a score \( A(s_i) \): 1 if fully correct, 0 if failed, and \( x \in (0,1) \) if partially correct. Task-level accuracy \( A(T) \) is computed as the mean of its subtasks, while overall accuracy \( A_{\text{overall}} \) is averaged over all evaluated tasks. For partial correctness (\( x \)), predefined rules are used to assign values based on specific evaluation criteria. This structured human evaluation provides a robust performance measure across all tasks (see Appendix~\ref{accuracy_extra} for a detailed explanation of the evaluation process and the rules for assigning partial scores).
\vspace{-1em}

\paragraph{Human Evaluation vs. CLIP Scores}
While automatic metrics like CLIP similarity are common for image/text editing, we use human evaluation for complex, multi-step, multimodal tasks. CLIP often misses small but critical changes (e.g., missing bounding boxes) and struggles with semantic coherence in multimodal tasks or tasks with multiple valid outputs. Our evaluation of 50 tasks with intentional errors showed CLIP similarity scores (0.93-0.98) significantly higher than human accuracy (0.7-0.8), highlighting CLIP's limitations (Table~\ref{tab:clip_vs_human}).

\begin{table}[h]
\vspace{-1.5em}

    \centering
    \caption{Comparison of CLIP Similarity vs. Human Evaluation on 50 tasks to assess CLIP similarity against human judgments in multimodal and multi-step editing.}
    \label{tab:clip_vs_human}
    \resizebox{\columnwidth}{!}{%
        \begin{tabular}{lcc}
            \toprule
            Metric & CLIP Similarity Score & Human Evaluation Accuracy \\
            \midrule
            Average (50 Tasks) & 0.96 & 0.78 \\
            \bottomrule
        \end{tabular}
    }
    \vspace{-1.5em}

\end{table}

% \begin{figure*}[t]
%     \centering
%     \includegraphics[width=1\linewidth]{sec/combine.pdf}
%     \vspace{-2.5em}
%     \caption{Dataset distribution for image-only (left) and text+image tasks (center) and quality comparison of different methods (right). \ours{} excels in complex multimodal tasks, with the radar plot highlighting its scalability.}
%     \label{fig:dataset}
%     \vspace{-0.5em}
% \end{figure*}

\paragraph{CLIP in Feedback Loops vs. Dataset Evaluation}
CLIP is effective for real-time subtask validation, as each subtask is assessed in isolation. In object detection, for instance, it evaluates only the detected region against the expected label (e.g., `car' or `person'), ensuring accurate verification. However, for full-task evaluation, CLIP prioritizes global similarity, often missing localized errors, making it unreliable for holistic assessment but useful for individual subtasks.\looseness-1

\begin{table}[h]
\vspace{-1.2em}

    \centering
    \caption{Correlation Analysis of CLIP vs Human Evaluation on 40 tasks, which indicates that human evaluation is still necessary. \looseness-1
    % comparing similarity scores with human evaluations for multimodal and multi-step tasks.
    }
    \label{tab:correlation_results}
    \begin{tabular}{lcc}
        \toprule
        Metric & Correlation Coefficient & $p$-value \\
        \midrule
        Spearman's $\rho$ & 0.59 & $6.07 \times 10^{-5}$ \\
        Kendall's $\tau$ & 0.47 & $5.83 \times 10^{-5}$ \\
        \bottomrule
    \end{tabular}
    \vspace{-1.5em}

\end{table}

\vspace{-0.5em}
\paragraph{Correlation Analysis}

We analyzed the correlation between CLIP scores and human accuracy across 40 tasks, finding weak agreement (Spearman’s \( \rho = 0.59 \), Kendall’s \( \tau = 0.47 \)). The low correlation confirms CLIP’s inability to capture nuanced inaccuracies, as visualized in Table \ref{tab:correlation_results} and the scatter plot in Appendix~\ref{correlation}.
\vspace{-1em}

% \paragraph{Accuracy Calculation} \label{accuracy}

% To quantify task performance, we manually assess subtask correctness using the following formulation:

% \begin{itemize}
%     \item \textbf{Subtask-Level Accuracy:} Each subtask \( s_i \) in a task \( T \) is assigned a score \( A(s_i) \), where:
%     \[
%     A(s_i) =
%     \begin{cases}
%     1, & \text{if fully correct} \\
%     0, & \text{if failed} \\
%     x \in (0,1), & \text{if partially correct}
%     \end{cases}
%     \]
%     \item \textbf{Task-Level Accuracy:} Computed as the mean of its subtasks:
%     \[
%     A(T) = \frac{1}{n} \sum_{i=1}^{n} A(s_i)
%     \]
%     \item \textbf{Overall Accuracy:} Averaged over \( m \) evaluated tasks:
%     \[
%     A_{\text{overall}} = \frac{1}{m} \sum_{j=1}^{m} A(T_j)
%     \]
% \end{itemize}

% \paragraph{Accuracy Calculation} \label{accuracy}

% Subtask correctness is manually assessed. Each subtask \( s_i \) in task \( T \) is assigned a score \( A(s_i) \): 1 if fully correct, 0 if failed, and \( x \in (0,1) \) if partially correct. Task-level accuracy \( A(T) \) is the mean of its subtasks. Overall accuracy \( A_{\text{overall}} \) is averaged over all evaluated tasks. This structured approach ensures accurate performance measurement across all tasks (see Appendix ~\ref{accuracy_extra} for a detailed explanation of the evaluation process).
% \vspace{-1em}

\paragraph{Execution Cost (Time)}

The cumulative execution time, including feedback-based retries and exploration of alternate models, is used to evaluate \ours{}’s efficiency.

\vspace{-1em}

\subsection{Main Results} \label{sec:main_results}
\vspace{-0.5em}

\paragraph{Performance Analysis} 

Table~\ref{tab:accuracy_comparison} demonstrates that \ours{} consistently outperforms baselines across all task categories. For simpler image-only tasks (1–2 subtasks), \ours{} achieves comparable accuracy, but as complexity increases (5+ subtasks), it significantly outperforms baselines. This is due to its A* search integration, which effectively refines LLM-generated plans, whereas baselines struggle with intricate workflows.

In text+image tasks, \ours{} achieves much higher accuracy due to its extensive toolset for text manipulation. Baselines, limited in tool variety, fail to perform well in multimodal scenarios. Additionally, \ours{}'s dynamic feedback and retry mechanisms further enhance robustness across diverse tasks, maintaining high-quality outputs. These results highlight its superiority in balancing cost and quality over agentic and non-agentic baselines.

% \begin{figure}[t]
%     \centering
%     \includegraphics[width=\linewidth]{sec/Quality Comparison.pdf}
%     \caption{Quality Comparison of Different Methods}
%     \label{fig:radar_quality}
% \end{figure}

% \begin{figure}
%     \centering
%     \includegraphics[width=\linewidth]{sec/Pareto_Front_Broken_Axis.pdf}
%     \caption{Pareto Chart Analysis}
%     \label{fig:pareto}
% \end{figure}
\vspace{-1em}

\paragraph{Radar Plot Analysis} 

Figure~\ref{fig:dataset} compares \ours{} with baselines across task complexities. While it shows marginal improvement in simple tasks, its advantage becomes pronounced in complex tasks (3+ subtasks), attributed to its path optimization and feedback integration. The radar plot confirms \ours{}'s scalability and multimodal capabilities, handling both image-only and text+image tasks effectively.
\vspace{-1em}

\paragraph{Pareto Optimality Analysis} 

% The Pareto front (Figure~\ref{fig:pareto}) highlights \ours{}'s ability to dynamically balance cost and quality by adjusting \( \alpha \). For \(\alpha = 0\), it prioritizes cost, achieving the lowest execution time while maintaining competitive quality. For \(\alpha = 2\), quality is maximized, albeit with slightly higher cost. Baselines fail to achieve this flexibility, further showcasing \ours{}'s superior optimization in cost-quality tradeoff. Baselines like CLOVA and GenArtist fall short of the Pareto front, as they fail to match \ours{}’s quality at comparable or lower costs, highlighting its superior optimization and execution.

The Pareto front (Figure~\ref{fig:pareto}) shows \ours{}'s ability to balance cost and quality by adjusting \( \alpha \). \(\alpha = 2\) prioritizes cost, while \(\alpha = 0\) maximizes quality. Baselines lack this flexibility and fall short of the Pareto front due to lower quality at comparable costs, demonstrating \ours{}’s superior cost-quality optimization.

% \begin{figure*}[t]
%     \centering
%     \includegraphics[width=\textwidth]{sec/qual.pdf}
%     \caption{Qualitative Examples}
%     \label{fig:qual_example}
% \end{figure*}

\renewcommand{\arraystretch}{1.2}
% \vspace{0em}
\vspace{-0.5em}
\begin{table}[h]
\vspace{-1em}
    \centering
    \caption{Comparison of key features across methods, highlighting the extensive set of capabilities supported by \ours{}, which are absent in baselines and contribute to its superior performance.}
    \label{tab:feature_support}
    \resizebox{\columnwidth}{!}{%
    \begin{tabular}{p{0.5\columnwidth}cccccc}
        \toprule
        \textbf{Feature} & \textbf{\ours{}} & \textbf{CLOVA} & \textbf{GenArtist} & \textbf{VisProg} & \textbf{Instruct Pix2Pix} \\
        \midrule
        Integration of LLM with A* Path Optimization & \checkmark & \texttimes & \texttimes & \texttimes & \texttimes \\
        Automatic Feedback Integration (Real-Time) & \checkmark & \texttimes & \texttimes & \texttimes & \texttimes \\
        Real-Time Cost-Quality Tradeoff & \checkmark & \texttimes & \texttimes & \texttimes & \texttimes \\
        User-Defined Cost-Quality Weightage & \checkmark & \texttimes & \texttimes & \texttimes & \texttimes \\
        Multimodality Support (Image + Text) & \checkmark & \texttimes & \texttimes & \texttimes & \texttimes \\
        Number of Tools for Task Accomplishment & 24 & \textless{}10 & \textless{}10 & \textless{}12 & \textless{}5 \\
        Feedback-Based Retrying and Model Selection & \checkmark & \checkmark & \checkmark & \texttimes & \texttimes \\
        Dynamic Adjustment of Heuristic Values & \checkmark & \texttimes & \texttimes & \texttimes & \texttimes \\
        \bottomrule
    \end{tabular}%
    }
    % \vspace{-1em}

\end{table}

\paragraph{Qualitative Results} 

Figure~\ref{fig:qual_comp} provides qualitative comparisons, illustrating \ours{}'s ability to seamlessly handle multimodal tasks. Table~\ref{tab:feature_support} highlights its distinct advantages, including real-time feedback, dynamic heuristic adjustments, and LLM integration with A* search—features lacking in baselines.
\vspace{-1em}

\paragraph{Summary}

\ours{} consistently outperforms baselines by integrating A* search, cost-quality optimization, and multimodal capabilities. It efficiently balances execution time and accuracy across a broader range of tasks, making it a highly adaptable solution for complex image and text-in-image editing tasks.

\vspace{-1em}

\subsection{Ablation Study} \label{sec:ablation_study}
\vspace{-0.5em}

We evaluate the impact of CoSTA*’s key components—\textbf{real-time feedback integration} and \textbf{multimodality support}—through targeted experiments, demonstrating their contributions to accuracy and task execution quality.

\begin{figure}
    \centering
    \includegraphics[width=0.8\linewidth]{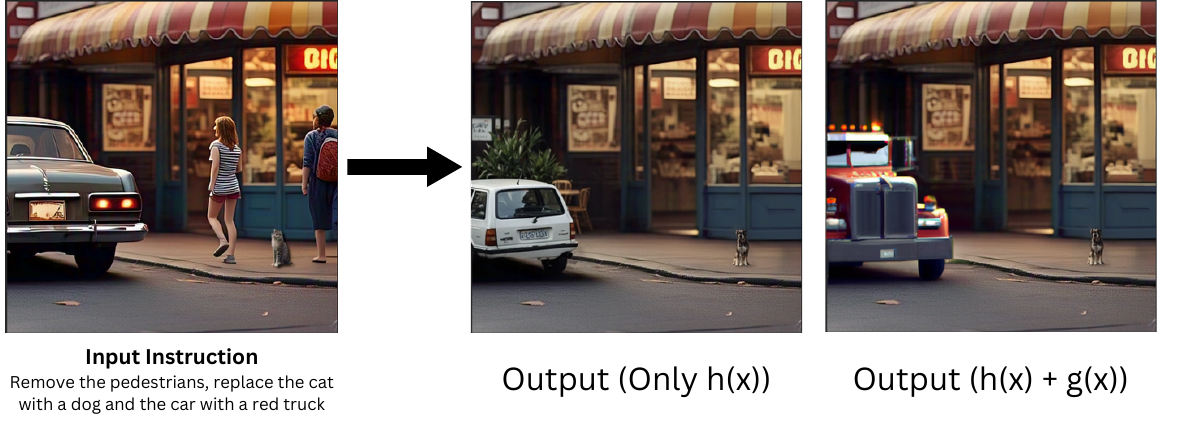}
    \vspace{-0.5em}

    \caption{Comparison of a task with \( h(x) \) and \( h(x) + g(x) \), showing how real-time feedback improves path selection and execution.}
    \label{fig:gx_hx_comparison}
    \vspace{-1em}

\end{figure}
\begin{table}[h]
\vspace{-1em}

    \centering
    \caption{Comparison of accuracy with and without \( g(x) \) on 35 high-risk tasks to analyze the impact of real-time feedback $g(x)$.}
    \label{tab:gx_vs_hx}
    \resizebox{0.6\columnwidth}{!}{%
        \begin{tabular}{lc}
            \toprule
            Approach & Average Accuracy \\
            \midrule
            \( h(x) \) Only & 0.798 \\
            \( h(x) + g(x)\) & 0.923 \\
            \bottomrule
        \end{tabular}
    }
    \vspace{-1em}

\end{table}
% \vspace{-1em}

\paragraph{Feedback Integration with \( g(x) \)}
To assess the role of real-time feedback, we compared two configurations across 35 high-risk tasks: one using only heuristic values (\( h(x) \)), and another integrating real-time feedback (\( g(x) \)). Tasks involved cases where tools with favorable \( h(x) \) sometimes underperformed, while alternatives with poorer heuristic scores excelled. As shown in Table~\ref{tab:gx_vs_hx}, using only \( h(x) \) resulted in 0.798 accuracy, whereas incorporating \( g(x) \) improved accuracy to 0.923 by penalizing poorly performing tools and dynamically selecting better alternatives. Figure~\ref{fig:gx_hx_comparison} illustrates an object replacement task where the \( h(x) \)-only approach failed but was corrected using \( g(x) \). This confirms that real-time feedback significantly enhances path selection and execution robustness.

\begin{table}[h]
    \vspace{-1em}

    \centering
    \caption{Comparison of image editing tools vs. \ours{} for text-based tasks. \ours{} outperforms image-only tools.}
    \label{tab:text_vs_image}
    \resizebox{\columnwidth}{!}{%
        \begin{tabular}{lcc}
            \toprule
            Metric & Image Editing Tools & CoSTA* \\
            \midrule
            Average Accuracy (30 Tasks) & 0.48 & 0.93 \\
            \bottomrule
        \end{tabular}
    }
    \vspace{-1em}
\end{table}

% \begin{figure}
%     \centering
%     \includegraphics[width=0.8\linewidth]{sec/Ablation.pdf}
%     \caption{Comparison of a Task with \( h(x) \) and \( h(x) + g(x) \), showing how real-time feedback improves path selection and execution.}
%     \label{fig:gx_hx_comparison}
% \end{figure}

\vspace{-1em}

\begin{figure}[h]
    \centering
    \includegraphics[width=0.8\linewidth]{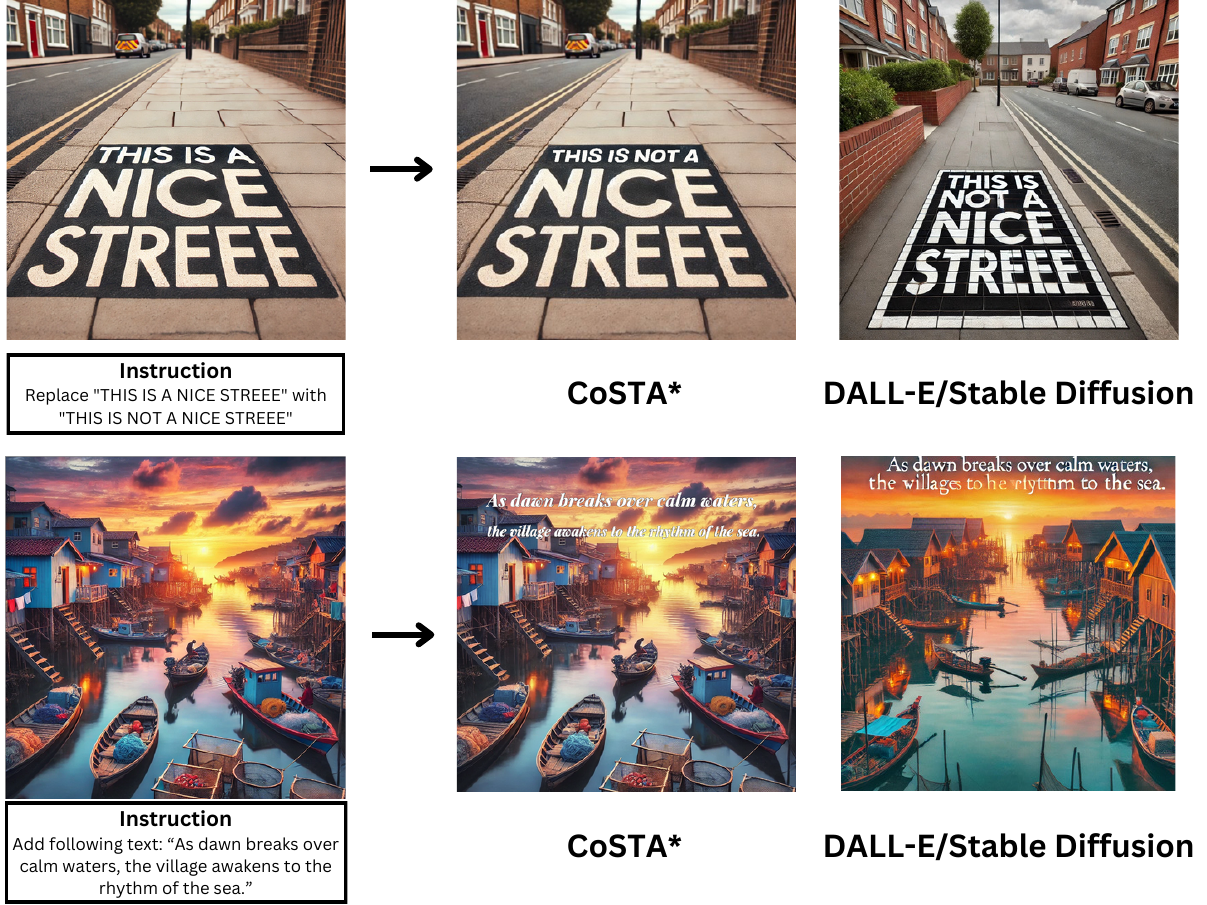}
    \caption{Qualitative comparison of image editing tools vs. \ours{} for text-based tasks, highlighting the advantages of our multimodal support in preserving visual and textual fidelity.}
    \label{fig:multimodality_impact}
\end{figure}

\paragraph{Impact of Multimodality Support}
We tested CoSTA* on tasks requiring both text and image processing, comparing performance against tools designed primarily for image modality like DALL-E\cite{DBLP:journals/corr/abs-2102-12092} and Stable Diffusion Inpaint\cite{DBLP:conf/cvpr/RombachBLEO22}. As shown in Table~\ref{tab:text_vs_image}, these tools struggled with text-related edits, achieving only 0.48 accuracy, while CoSTA* retained visual and textual fidelity, reaching 0.93 accuracy. Figure~\ref{fig:multimodality_impact} highlights the qualitative advantages of multimodal support, reinforcing that integrating specialized models for text manipulation leads to significantly better results.

\vspace{-1.2em}
\section{Conclusions} \label{sec:conclusions}
\vspace{-0.5em}
In this paper, we present a novel image editing agent that leverages the capabilities of a large multimodal model as a planner combined with the flexibility of the A* algorithm to search for an optimal editing path, balancing the cost-quality tradeoff. Experimental results demonstrate that \ours effectively handles complex, real-world editing queries with reliability while surpassing existing baselines in terms of image quality. Moreover, our proposed agent supports 24 tasks, significantly more than the current state-of-the-art. We believe that this neurosymbolic approach is a promising direction toward more capable and reliable agents in the future.
\vspace{-0.5em}

% \begin{figure}[h]
%     \centering
%     \includegraphics[width=0.8\linewidth]{sec/multimodality_impact.pdf}
%     \caption{Qualitative comparison of image editing tools vs. \ours{} for text-based tasks, highlighting the advantages of our multimodal support in preserving visual and textual fidelity.}
%     \label{fig:multimodality_impact}
% \end{figure}

\section*{Impact Statement}
This paper aims to advance the field of multimodal machine learning by improving image and text-in-image editing through optimized task decomposition and execution. Our work enhances automation in content creation, accessibility, and image restoration, contributing positively to various applications. However, as with any image manipulation system, there is a potential risk of misuse, such as generating misleading content or altering visual information in ways that could contribute to misinformation. While our approach itself does not promote unethical use, we acknowledge the importance of responsible deployment and advocate for safeguards such as provenance tracking and watermarking to ensure transparency. Additionally, since our method relies on pre-trained models, inherent biases in those models may persist. Addressing fairness through dataset audits and bias mitigation remains an important consideration for future research. Overall, our work strengthens AI-driven editing capabilities while emphasizing the need for ethical and responsible usage.

% \begin{figure*}[]
%     \centering
%     \includegraphics[width=1\textwidth]{sec/Qualitative_Comparison.pdf}
%     \caption{Qualitative Comparison with State-of-the-Art Image Editing Models. We compare \ours{} with state-of-the-art image editing models, including GenArtist\cite{wang2024genartistmultimodalllmagent}, MagicBrush\cite{zhang2024magicbrushmanuallyannotateddataset}, InstructPix2Pix\cite{brooks2023instructpix2pixlearningfollowimage}, and CLOVA\cite{DBLP:conf/cvpr/Gao0ZMHZL24}. The input images and prompts are shown alongside the outputs generated by each method, illustrating differences in accuracy, visual coherence, and the ability to handle complex multimodal tasks. Step-by-step execution of some tasks using \ours{}, detailing intermediate subtask outputs, is presented in Figure~\ref{fig:qual_example}.}
%     \label{fig:qual_comp}
% \end{figure*}

% \begin{figure*}[t]
%     \centering
%     \includegraphics[width=\textwidth]{sec/qual.pdf}
%     \caption{Qualitative Examples}
%     \label{fig:qual_example}
% \end{figure*}
\nocite{langley00}

\bibliography{CoSTA/main}
\bibliographystyle{icml2025}

%%%%%%%%%%%%%%%%%%%%%%%%%%%%%%%%%%%%%%%%%%%%%%%%%%%%%%%%%%%%%%%%%%%%%%%%%%%%%%%
%%%%%%%%%%%%%%%%%%%%%%%%%%%%%%%%%%%%%%%%%%%%%%%%%%%%%%%%%%%%%%%%%%%%%%%%%%%%%%%
% APPENDIX
%%%%%%%%%%%%%%%%%%%%%%%%%%%%%%%%%%%%%%%%%%%%%%%%%%%%%%%%%%%%%%%%%%%%%%%%%%%%%%%
%%%%%%%%%%%%%%%%%%%%%%%%%%%%%%%%%%%%%%%%%%%%%%%%%%%%%%%%%%%%%%%%%%%%%%%%%%%%%%%
\newpage
\appendix
\onecolumn

\begin{figure*}[t]
    \centering
    \includegraphics[width=\textwidth]{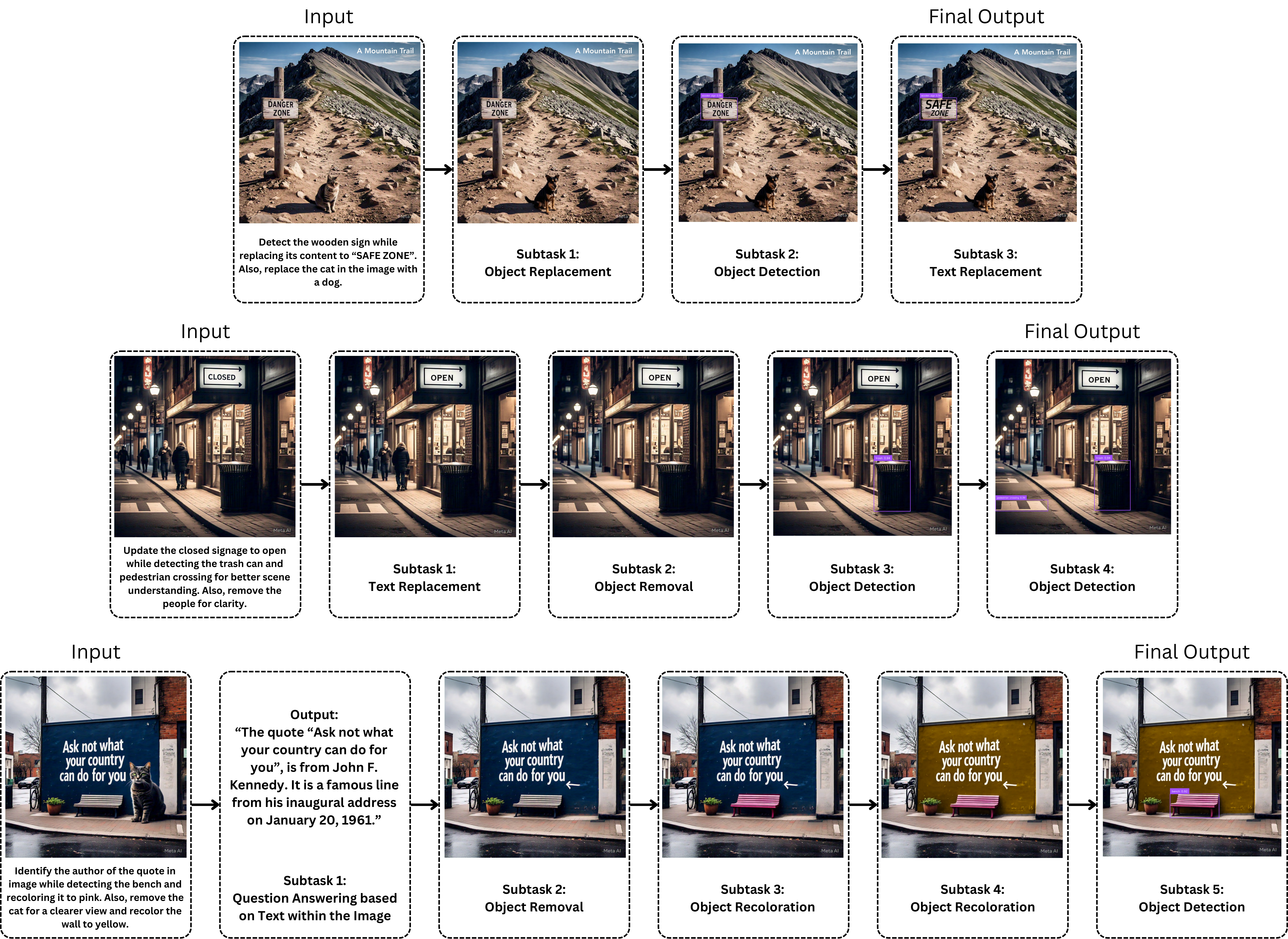}
    \caption{Step-by-step execution of editing tasks using \ours{}. Each row illustrates an input image, the corresponding subtask breakdown, and intermediate outputs at different stages of the editing process. This visualization highlights how \ours{} systematically refines outputs by leveraging specialized models for each subtask, ensuring greater accuracy and consistency in multimodal tasks.}
    \label{fig:qual_example}
    \vspace{-2em}
\end{figure*}
% \vspace{-5em}
\section{Step-by-Step Execution of Tasks in Figure \ref{fig:qual_comp}}
\label{sec:step_by_step_execution}

To complement the qualitative comparisons presented in Figure \ref{fig:qual_comp}, Figure~\ref{fig:qual_example} provides a visualization of the step-by-step execution of selected subtasks within the composite task by \ours{}. This figure highlights the intermediate outputs produced by each subtask, illustrating how complex image editing operations are decomposed and executed sequentially.

By showcasing the incremental progression of subtasks, this visualization provides a clearer view of how different intermediate outputs contribute to the final edited image. Rather than illustrating the full decision-making process of \ours{}, the figure focuses on the stepwise transformations applied to the image, offering a practical reference for understanding the effects of each subtask.

This breakdown highlights key transitions in tasks, demonstrating the intermediate results generated at various stages. It provides insight into how each operation modifies the image, helping to better interpret the qualitative comparisons presented in the main text.

\vspace{-1em}

\section{Human Evaluation for Accuracy Calculation} \label{accuracy_extra}
\vspace{-0.7em}

To ensure reliable performance assessment, we conduct human evaluations for accuracy calculation across all subtasks and tasks. Unlike automatic metrics such as CLIP similarity, human evaluation accounts for nuanced errors, semantic inconsistencies, and multi-step dependencies that are often missed by automated tools. This section outlines the evaluation methodology, scoring criteria, and aggregation process.

\begin{table}[h]
    \centering
    \caption{Predefined Rules for Assigning Partial Correctness Scores in Human Evaluation}
    \label{tab:partial_correctness}
    \resizebox{0.7\columnwidth}{!}{%
    \begin{tabular}{l p{9cm} c}
        \toprule
        \textbf{Task Type} & \textbf{Evaluation Criteria} & \textbf{Assigned Score} \\
        \midrule
        \multirow{6}{*}{Image-Only Tasks} 
        & Minor artifacts, barely noticeable distortions & 0.9 \\
        & Some visible artifacts, but main content is unaffected & 0.8 \\
        & Noticeable distortions, but retains basic correctness & 0.7 \\
        & Significant artifacts or blending issues & 0.5 \\
        & Major distortions or loss of key content & 0.3 \\
        & Output is almost unusable, but some attempt is visible & 0.1 \\
        \midrule
        \multirow{6}{*}{Text+Image Tasks} 
        & Text is correctly placed but slightly misaligned & 0.9 \\
        & Font or color inconsistencies, but legible & 0.8 \\
        & Noticeable alignment or formatting issues & 0.7 \\
        & Some missing or incorrect words but mostly readable & 0.5 \\
        & Major formatting errors or loss of intended meaning & 0.3 \\
        & Text placement is incorrect, missing, or unreadable & 0.1 \\
        \bottomrule
    \end{tabular}
    }
\end{table}

\subsection{Subtask-Level Accuracy}

Each subtask \( s_i \) in a task \( T \) is manually assessed by evaluators and assigned a correctness score \( A(s_i) \) based on the following criteria:

\begin{equation}
A(s_i) =
\begin{cases}
1, & \text{if the subtask is fully correct} \\
x, & \text{if the subtask is partially correct, where } x \in \{0.1, 0.3, 0.5, 0.7, 0.8, 0.9\} \\
0, & \text{if the subtask has failed}
\end{cases}
\end{equation}

Partial correctness (\( x \)) is determined based on predefined task-specific criteria. Table~\ref{tab:partial_correctness} defines the rules used to assign these scores across different subtasks. 

\subsection{Task-Level Accuracy}

Task accuracy is computed as the mean correctness of its subtasks:

\begin{equation}
A(T) = \frac{1}{|S_T|} \sum_{i=1}^{|S_T|} A(s_i)
\end{equation}

where \( S_T \) is the set of subtasks in task \( T \), ensuring that task accuracy reflects overall subtask correctness.

\subsection{Overall Accuracy Across Tasks}

To evaluate system-wide performance, the overall accuracy is computed as the average of task-level accuracies:

\begin{equation}
A_{\text{overall}} = \frac{1}{|T|} \sum_{j=1}^{|T|} A(T_j)
\end{equation}

where \( |T| \) is the total number of evaluated tasks.

\section{Automatic Construction of the Tool Dependency Graph}
\label{sec:tdg_auto}

The Tool Dependency Graph (TDG) can be automatically generated by analyzing the input-output relationships of each tool. Each tool $v_i$ is associated with a set of required inputs $\mathcal{I}(v_i)$ and a set of produced outputs $\mathcal{O}(v_i)$. We construct directed edges $(v_i, v_j)$ whenever $\mathcal{O}(v_i) \cap \mathcal{I}(v_j) \neq \emptyset$, meaning the output of tool $v_i$ is required as input for tool $v_j$.

These input-output relationships are explicitly listed in the \textbf{Model Description Table (MDT)}, where two dedicated columns specify the expected inputs and produced outputs for each tool. Using this structured metadata, the TDG can be dynamically constructed without manual intervention, ensuring that dependencies are correctly captured and automatically updated as the toolset evolves.
\vspace{-1em}
\section{Dataset Generation and Evaluation Setup} \label{Dataset}

\subsection{A. Dataset Construction for Benchmarking}

To rigorously evaluate the effectiveness of our method, we constructed a diverse, large-scale dataset designed to test various image editing tasks under complex, multi-step, and multimodal constraints. The dataset generation process was carefully structured to ensure both realism and consistency in task complexity.

\subsubsection{1. Automatic Prompt Generation \& Human Curation}
To simulate real-world image editing tasks, we first generated a diverse set of structured prompts using a \textbf{Large Language Model (LLM)}. These prompts were designed to cover a wide variety of editing operations, including:
\begin{itemize}
    \item Object \textbf{replacement, addition, removal, and recoloration},
    \item \textbf{Text-based modifications} such as replacement, addition, and redaction,
    \item \textbf{Scene-level changes}, including background modification and outpainting.
\end{itemize}

While LLM-generated prompts provided an automated way to scale dataset creation, they lacked real-world editing constraints. Thus, each prompt was manually curated by human annotators to ensure:
\begin{enumerate}
    \item \textbf{Logical Feasibility:} Ensuring that edits could be performed realistically on an image.
    \item \textbf{Complexity Diversity:} Creating \textbf{simple (1-2 subtasks)} and \textbf{complex (5+ subtasks)} tasks for a comprehensive evaluation.
    \item \textbf{Ensuring Clarity:} Refining ambiguous phrasing or vague instructions.
\end{enumerate}

\subsubsection{2. Image Generation with Meta AI}
Once the curated prompts were finalized, \textbf{image generation} was performed using \textbf{Meta AI’s generative model}. Unlike generic image generation, our \textbf{human annotators provided precise instructions} to ensure that:
\begin{itemize}
    \item \textbf{Every key element mentioned in the prompt} was included in the generated image.
    \item The \textbf{scene, object attributes, and text elements} were visually clear for the intended edits.
    \item The images had sufficient complexity and diversity to challenge different image-editing models.
\end{itemize}

For example, if a prompt requested \textit{``Replace the red bicycle with a blue motorcycle and remove the tree in the background,''} the generated image explicitly contained a \textbf{red bicycle and a clearly distinguishable tree}, ensuring that subsequent edits could be precisely evaluated.

\subsection{B. Dataset Composition \& Subtask Distribution}

Our dataset comprises \textbf{121 total image-task pairs}, with tasks spanning both \textbf{image-only} and \textbf{text+image} categories. Each image-editing prompt is decomposed into \textbf{subtasks}, which are then mapped to the supported models for evaluation.

\begin{figure}[ht]
% \vspace{-0em}
    \centering
    \includegraphics[width=0.7\textwidth]{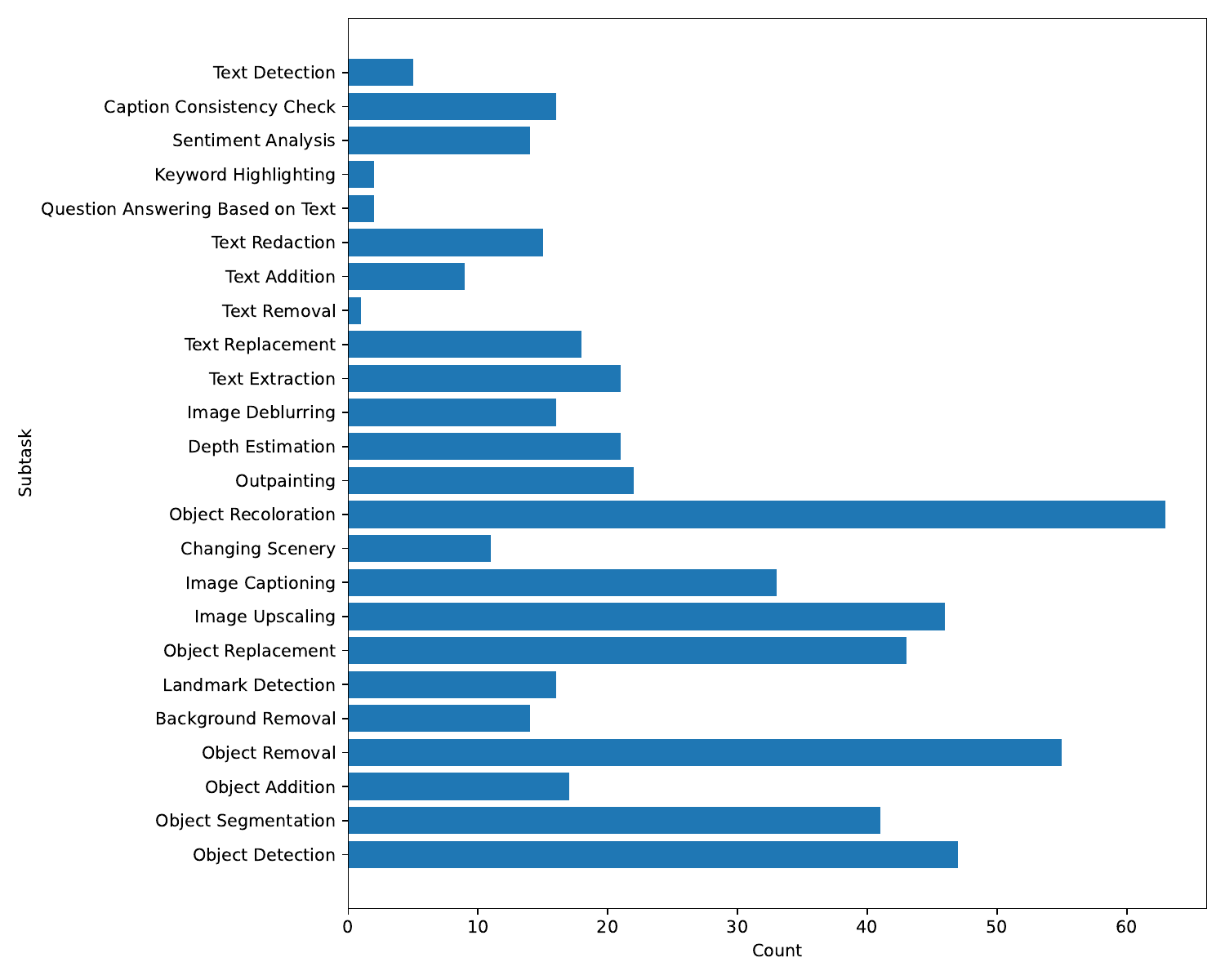}
    \caption{Distribution of the number of instances for each subtask in the dataset.}
    \label{fig:subtask_distribution}
    \vspace{-0em}
\end{figure}

Figure~\ref{fig:subtask_distribution} illustrates the distribution of subtasks across the dataset. This provides insights into:
\begin{itemize}
    \item The \textbf{relative frequency} of each subtask.
    \item The \textbf{balance between different categories} (e.g., object-based, text-based, scene-based).
\end{itemize}
\begin{figure}[H]
\vspace{-0em}
    \centering
    \includegraphics[width=0.7\textwidth]{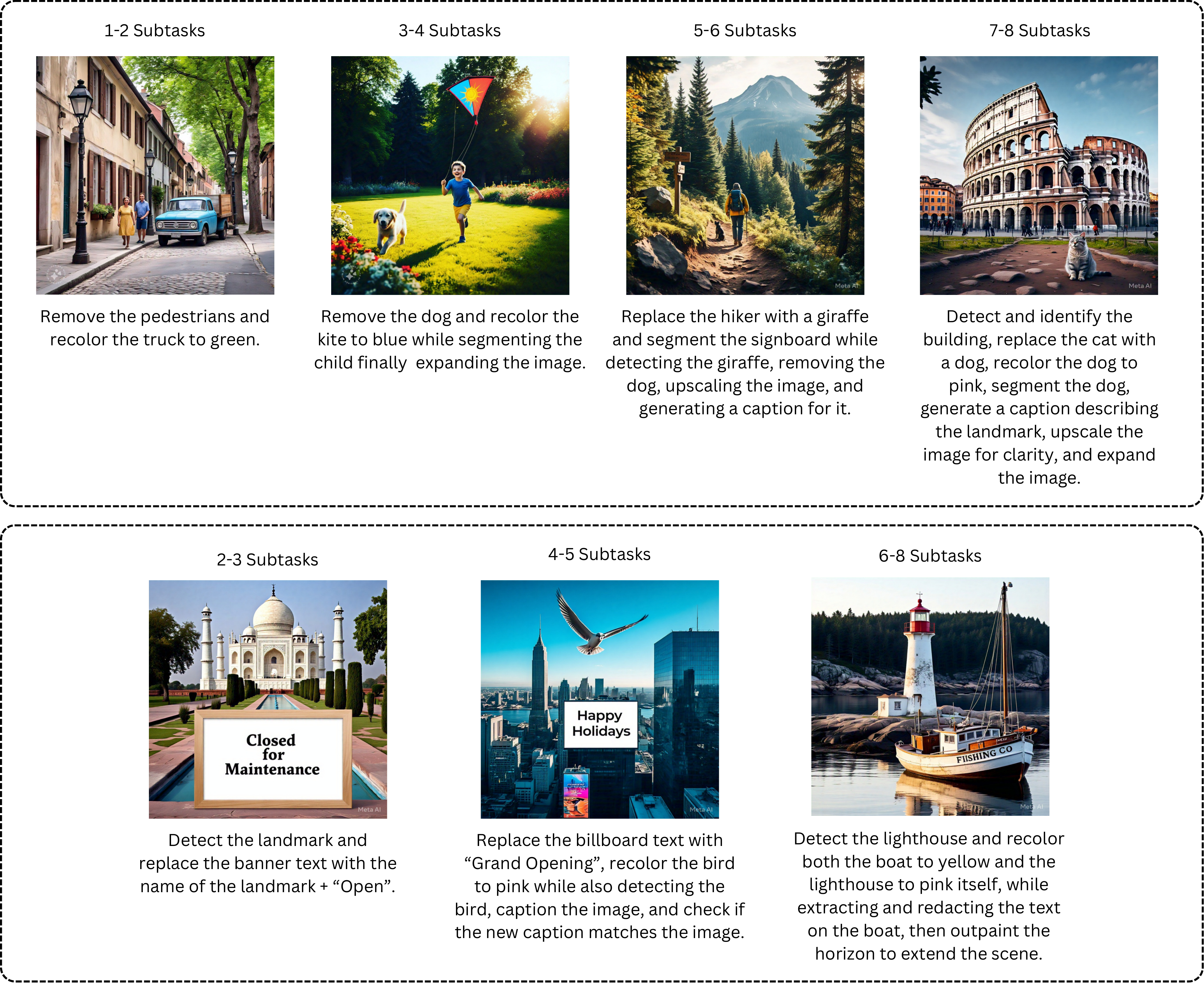}
    \caption{An overview of the dataset used for evaluation, showcasing representative input images and prompts across different task categories. The top section presents examples from image-only tasks, while the bottom section includes text+image tasks. These examples illustrate the diversity of tasks in our dataset, highlighting the range of modifications required for both visual and multimodal editing scenarios.}
    \label{fig:subtask_distribution}
    \vspace{-0em}
\end{figure}

The dataset ensures adequate representation of each subtask, avoiding skew toward a specific category. The most common subtasks in the dataset include \textbf{Object Replacement, Object Recoloration, and Object Removal}, while rarer but complex operations like \textbf{Keyword Highlighting} remain crucial for evaluation.

\begin{table}[h]
\vspace{-1em}
    \centering
    \caption{Average CLIP Similarity Scores for Outputs of Randomness-Prone Subtasks}
    \label{tab:consistency}
    \small % Reduce font size for compact table
    \setlength{\tabcolsep}{4pt} % Reduce horizontal padding
    \begin{tabular}{lc}
        \toprule
        \textbf{Subtask} & \textbf{Avg Similarity Score} \\
        \midrule
        Object Replacement       & 0.98  \\ 
        Object Recoloration      & 0.99  \\ 
        Object Addition          & 0.97  \\ 
        Object Removal           & 0.97 \\ 
        Image Captioning         & 0.92  \\ 
        Outpainting              & 0.99 \\ 
        Changing Scenery         & 0.96  \\ 
        Text Removal             & 0.98  \\ 
        QA on Text               & 0.96 \\ 
        \bottomrule
    \end{tabular}
    \vspace{-1em}
\end{table}
\FloatBarrier
\vspace{-1em}
\section{Consistency in CoSTA* Outputs} \label{sec:consistency}
% \vspace{-1em}
To assess robustness against randomness, we evaluated CoSTA* on subtasks prone to variability, such as object replacement and recoloration, where outputs may slightly differ across executions (e.g., different dog appearances when replacing a cat). A set of 20 images per subtask was selected, and each was processed multiple times. Outputs for each image were compared among each other using CLIP similarity scores, measuring consistency. As summarized in Table~\ref{tab:consistency}, CoSTA* maintains high similarity across runs, confirming its stability. Variability was negligible in most cases, except for image captioning (0.92 similarity), where multiple valid descriptions naturally exist. These results demonstrate that CoSTA* is highly consistent, with minimal impact from randomness.
\vspace{-1em}

\begin{table*}[h]
    \centering
    \caption{Model Description Table (MDT). Each model is listed with its supported subtasks, input dependencies, and outputs.}
    \label{tab:full_mdt}
    \resizebox{\textwidth}{!}{%
    \begin{tabular}{l p{5cm} p{5cm} p{5cm}}
        \toprule
        \textbf{Model} & \textbf{Tasks Supported} & \textbf{Inputs} & \textbf{Outputs} \\
        \midrule
        Grounding DINO\cite{liu2024groundingdinomarryingdino} & Object Detection & Input Image & Bounding Boxes \\
        YOLOv7\cite{wang2022yolov7trainablebagoffreebiessets} & Object Detection & Input Image & Bounding Boxes \\
        SAM\cite{kirillov2023segment} & Object Segmentation & Bounding Boxes & Segmentation Masks \\
        DALL-E\cite{DBLP:journals/corr/abs-2102-12092} & Object Replacement & Segmentation Masks & Edited Image \\
        DALL-E\cite{DBLP:journals/corr/abs-2102-12092} & Text Removal & Text Region Bounding Box & Image with Removed Text \\
        Stable Diffusion Erase\cite{rombach2022highresolutionimagesynthesislatent} & Text Removal & Text Region Bounding Box & Image with Removed Text \\
        Stable Diffusion Inpaint\cite{rombach2022highresolutionimagesynthesislatent} & Object Replacement, Object Recoloration, Object Removal & Segmentation Masks & Edited Image \\
        Stable Diffusion Erase\cite{rombach2022highresolutionimagesynthesislatent} & Object Removal & Segmentation Masks & Edited Image \\
        Stable Diffusion 3\cite{rombach2022highresolutionimagesynthesislatent} & Changing Scenery & Input Image & Edited Image \\
        Stable Diffusion Outpaint\cite{rombach2022highresolutionimagesynthesislatent} & Outpainting & Input Image & Expanded Image \\
        Stable Diffusion Search \& Recolor\cite{rombach2022highresolutionimagesynthesislatent} & Object Recoloration & Input Image & Recolored Image \\
        Stable Diffusion Remove Background\cite{rombach2022highresolutionimagesynthesislatent} & Background Removal & Input Image & Edited Image \\
        Text Removal (Painting) & Text Removal & Text Region Bounding Box & Image with Removed Text \\
        DeblurGAN\cite{kupyn2018deblurganblindmotiondeblurring} & Image Deblurring & Input Image & Deblurred Image \\
        LLM (GPT-4o) & Image Captioning & Input Image & Image Caption \\
         LLM (GPT-4o) & Question Answering based on text, Sentiment Analysis & Extracted Text, Font Style Label & New Text, Text Region Bounding Box, Text Sentiment, Answers to Questions \\
        Google Cloud Vision\cite{googlecloudvision} & Landmark Detection & Input Image & Landmark Label \\
        CRAFT\cite{baek2019characterregionawarenesstext} & Text Detection & Input Image & Text Bounding Box \\
        CLIP\cite{radford2021learningtransferablevisualmodels} & Caption Consistency Check & Extracted Text & Consistency Score \\
        DeepFont\cite{wang2015deepfontidentifyfontimage} & Text Style Detection & Text Bounding Box & Font Style Label \\
        EasyOCR\cite{rakpong_kittinaradorn_2022_6850706} & Text Extraction & Text Bounding Box & Extracted Text \\
        MagicBrush\cite{zhang2024magicbrushmanuallyannotateddataset} & Object Addition & Input Image & Edited Image with Object \\
        pix2pix\cite{isola2018imagetoimagetranslationconditionaladversarial} & Changing Scenery (Day2Night) & Input Image & Edited Image \\
        Real-ESRGAN\cite{DBLP:conf/iccvw/WangXDS21} & Image Upscaling & Input Image & High-Resolution Image \\
        Text Writing using Pillow (For Addition) & Text Addition & New Text, Text Region Bounding Box & Image with Text Added \\
        Text Writing using Pillow & Text Replacement, Keyword Highlighting & Image with Removed Text & Image with Text Added \\
        Text Redaction (Code-based) & Text Redaction & Text Region Bounding Box & Image with Redacted Text \\
        MiDaS & Depth Estimation & Input Image & Image with Depth of Objects \\
        \bottomrule
    \end{tabular}
    }
\end{table*}

\section{Model Description Table (MDT)}

The full Model Description Table (MDT) provides a comprehensive list of all 24 specialized models used in the \ours{} pipeline for image and text-in-image editing. Each model is mapped to its supported subtasks, input dependencies, and outputs, ensuring optimal tool selection for diverse editing requirements. These structured input-output relationships enable the automatic construction of the \textbf{Tool Dependency Graph (TDG)} by identifying dependencies between models based on their required inputs and generated outputs. The inputs and outputs are mentioned only in a way to design dependencies between models to construct the Tool Dependency Graph. Some standard inputs are skipped in this table for simplicity (e.g. each model's input would also include the previous model's output image, some models might also require extra inputs which were outputted not by the previous model but by a model back in the path, etc.). Unlike generic pipelines, \ours{} utilizes targeted models to enhance accuracy and efficiency in text-related visual tasks. Table~\ref{tab:full_mdt} presents the complete MDT, detailing the capabilities of each model across different task categories and their role in facilitating automated dependency resolution.
\vspace{-1em}

\section{Benchmark Table (BT)}

The Benchmark Table (BT) defines execution time and accuracy scores for each tool-task pair \( BT(v_i, s_j) \), where \( v_i \) is a tool and \( s_j \) is a subtask. It serves as a baseline for A$^*$ search, enabling efficient tool selection. Both execution time and accuracy scores are based on empirical evaluations and published benchmarks (wherever available). For tools without prior benchmarks, evaluations on 137 instances of the specific subtask were conducted on 121 images from the dataset, with results included in Table~\ref{tab:bt}. Accuracy values are normalized with respect to max within each subtask on a [0,1] scale for comparability.

\begin{table*}[h]
    \centering
\caption{Benchmark Table for Accuracy and Execution Time. Accuracy and execution time for each tool-task pair are obtained from cited sources where available. For tools without prior benchmarks, evaluation was conducted over \textbf{137 instances of the specific subtask on 121 images from the dataset}, ensuring a robust assessment across varied conditions.}
\vspace{1em}
    \label{tab:bt}
    \resizebox{\textwidth}{!}{%
    \begin{tabular}{l l c c | l}
        \toprule
\textbf{Model Name} & \textbf{Subtask} & \textbf{Accuracy} & \textbf{Time (s)} & \textbf{Source} \\
\midrule
        DeblurGAN\cite{kupyn2018deblurganblindmotiondeblurring} & Image Deblurring & 1.00 & 0.8500 & \cite{kupyn2018deblurganblindmotiondeblurring} \\
        MiDaS\cite{ranftl2020robustmonoculardepthestimation} & Depth Estimation & 1.00 & 0.7100 & Evaluation on 137 instances of this subtask \\
        YOLOv7\cite{wang2022yolov7trainablebagoffreebiessets} & Object Detection & 0.82 & 0.0062 & \cite{wang2022yolov7trainablebagoffreebiessets} \\
        Grounding DINO\cite{liu2024groundingdinomarryingdino} & Object Detection & 1.00 & 0.1190 & Accuracy: \cite{liu2024groundingdinomarryingdino}, Time: Evaluation on 137 instances of this subtask \\
        CLIP\cite{radford2021learningtransferablevisualmodels} & Caption Consistency Check & 1.00 & 0.0007 & Evaluation on 137 instances of this subtask \\
        SAM\cite{ravi2024sam2segmentimages} & Object Segmentation & 1.00 & 0.0460 & Accuracy: Evaluation on 137 instances of this subtask, Time: \cite{ravi2024sam2segmentimages} \\
        CRAFT\cite{baek2019characterregionawarenesstext} & Text Detection & 1.00 & 1.2700 & Accuracy: \cite{baek2019characterregionawarenesstext}, Time: Evaluation on 137 instances of this subtask \\
        Google Cloud Vision\cite{googlecloudvision} & Landmark Detection & 1.00 & 1.2000 & Evaluation on 137 instances of this subtask \\
        EasyOCR\cite{rakpong_kittinaradorn_2022_6850706} & Text Extraction & 1.00 & 0.1500 & Evaluation on 137 instances of this subtask \\
        Stable Diffusion Erase\cite{DBLP:conf/cvpr/RombachBLEO22} & Object Removal & 1.00 & 13.8000 & Evaluation on 137 instances of this subtask \\
        DALL-E\cite{DBLP:journals/corr/abs-2102-12092} & Object Replacement & 1.00 & 14.1000 & Evaluation on 137 instances of this subtask \\
        Stable Diffusion Inpaint\cite{DBLP:conf/cvpr/RombachBLEO22} & Object Removal & 0.93 & 12.1000 & Evaluation on 137 instances of this subtask \\
        Stable Diffusion Inpaint\cite{DBLP:conf/cvpr/RombachBLEO22} & Object Replacement & 0.97 & 12.1000 & Evaluation on 137 instances of this subtask \\
        Stable Diffusion Inpaint\cite{DBLP:conf/cvpr/RombachBLEO22} & Object Recoloration & 0.89 & 12.1000 & Evaluation on 137 instances of this subtask \\
        Stable Diffusion Search \& Recolor\cite{DBLP:conf/cvpr/RombachBLEO22} & Object Recoloration & 1.00 & 14.7000 & Evaluation on 137 instances of this subtask \\
        Stable Diffusion Outpaint\cite{DBLP:conf/cvpr/RombachBLEO22} & Outpainting & 1.00 & 12.7000 & Evaluation on 137 instances of this subtask \\
        Stable Diffusion Remove Background\cite{DBLP:conf/cvpr/RombachBLEO22} & Background Removal & 1.00 & 12.5000 & Evaluation on 137 instances of this subtask \\
        Stable Diffusion 3\cite{DBLP:conf/cvpr/RombachBLEO22} & Changing Scenery & 1.00 & 12.9000 & Evaluation on 137 instances of this subtask \\
        pix2pix\cite{isola2018imagetoimagetranslationconditionaladversarial} & Changing Scenery (Day2Night) & 1.00 & 0.7000 & Accuracy: \cite{isola2018imagetoimagetranslationconditionaladversarial}, Time: Evaluation on 137 instances of this subtask \\
        Real-ESRGAN\cite{DBLP:conf/iccvw/WangXDS21} & Image Upscaling & 1.00 & 1.7000 & Evaluation on 137 instances of this subtask \\
        LLM (GPT-4o) & Question Answering based on Text & 1.00 & 6.2000 & Evaluation on 137 instances of this subtask \\
        LLM (GPT-4o) & Sentiment Analysis & 1.00 & 6.1500 & Evaluation on 137 instances of this subtask \\
        LLM (GPT-4o) & Image Captioning & 1.00 & 6.3100 & Evaluation on 137 instances of this subtask \\
        DeepFont\cite{wang2015deepfontidentifyfontimage} & Text Style Detection & 1.00 & 1.8000 & Evaluation on 137 instances of this subtask \\
        Text Writing - Pillow & Text Replacement & 1.00 & 0.0380 & Evaluation on 137 instances of this subtask \\

        Text Writing - Pillow & Text Addition & 1.00 & 0.0380 & Evaluation on 137 instances of this subtask \\

        Text Writing - Pillow & Keyword Highlighting & 1.00 & 0.0380 & Evaluation on 137 instances of this subtask \\
        MagicBrush\cite{DBLP:conf/nips/ZhangMCSS23} & Object Addition & 1.00 & 12.8000 & Accuracy: \cite{DBLP:conf/nips/ZhangMCSS23}, Time: Evaluation on 137 instances of this subtask \\
        Text Redaction & Text Redaction & 1.00 & 0.0410 & Evaluation on 137 instances of this subtask \\
        Text Removal by Painting & Text Removal (Fallback) & 0.20 & 0.0450 & Evaluation on 137 instances of this subtask \\
        DALL-E\cite{DBLP:journals/corr/abs-2102-12092} & Text Removal & 1.00 & 14.2000 & Evaluation on 137 instances of this subtask \\
        Stable Diffusion Erase\cite{DBLP:conf/cvpr/RombachBLEO22} & Text Removal & 0.97 & 13.8000 & Evaluation on 137 instances of this subtask \\
        \bottomrule
    \end{tabular}
    }
\end{table*}

\section{Algorithms}
% \vspace{-1em}
\begin{algorithm}[H]
\caption{A* Search for Optimal Toolpath}\label{alg:a-star-search}
\KwIn{Tool Subgraph $G_{ts}$, Benchmark Table $BT$, Tradeoff Parameter $\alpha$, Quality Threshold}
\KwOut{Optimal Execution Path}

\SetKwFunction{Neighbors}{Neighbors}
\SetKwFunction{CalculateHeuristic}{CalculateHeuristic}
\SetKwFunction{ComputeExecutionCost}{ComputeExecutionCost}
\SetKwFunction{CalculateActualCost}{CalculateActualCost}
\SetKwFunction{CalculateActualQuality}{CalculateActualQuality}
\SetKwFunction{QualityCheck}{QualityCheck}
\SetKwFunction{RetryMechanism}{RetryMechanism}
\SetKwFunction{Min}{Min}

\textbf{Step 1: Initialize Search} \\
Initialize Priority Queue $Q$\;
Initialize $g(x) \gets \infty$ for all nodes except root\;

Precompute heuristic values for all nodes:  
\ForEach{$v$ in $G_{ts}$}{
    $h(v) \gets$ \CalculateHeuristic{$BT$, $v$, $\alpha$}\;
}

Initialize Start Node:  
Set Input Image as Root Node $r$\;
$g(r) \gets 0$\;
$f(r) \gets h(r)$\;
Push $(f(r), [r])$ into $Q$\;
Mark $r$ as Open\;

\While{$Q$ is not empty}{
    $(f(x), \text{current\_path}) \gets \text{Pop}(Q)$\;
    $x \gets \text{LastNode}(\text{current\_path})$\;

    \If{$x$ is a leaf node}{
        \Return{$\text{current\_path}$}
    }

    \ForEach{neighbor $y$ in \Neighbors{$x$}}{

        $c(y) \gets$ \CalculateActualCost{$y$}\;
        $q(y) \gets$ \CalculateActualQuality{$y$}\;
        $g(y)_{\text{new}} \gets$ \ComputeExecutionCost{$g(x)$, $c(y)$, $q(y)$, $\alpha$}\;

        \If{\QualityCheck{$y$} $\geq$ Quality Threshold}{
            $g(y) \gets$ \Min{$g(y)_{\text{new}}$, $g(y)$}\;
        }
        \Else{
            $g(y)_{\text{new2}} \gets$ \RetryMechanism{$y$}\;
            \If{\QualityCheck{$y$} $\geq$ Quality Threshold}{
                $g(y)_{\text{final}} \gets g(y)_{\text{new}} + g(y)_{\text{new2}}$\;
                $g(y) \gets$ \Min{$g(y)_{\text{final}}$, $g(y)$}\;
            }
            \Else{
                \textbf{continue};  Node remains unexplored
            }
        }

        $f(y) \gets g(y) + h(y)$\;

        Push $(f(y), \text{current\_path} + [y])$ into $Q$\;
    }
    
}

\textbf{Step 2: Output Optimal Path} \\
Terminate when the lowest-cost valid path is found\;
\Return{Optimal Path}\;
\end{algorithm}
% \vspace{-5em}

\begin{algorithm}[H]
\caption{Tool Subgraph Construction}\label{alg:toolpath-graph-construction}
\KwIn{Image $x$, Prompt $u$, Tool Dependency Graph $G_{td}$, Model Description Table $MDT$, Supported Subtasks $\mathcal{S}$}
\KwOut{Tool Subgraph $G_{ts}$}

\SetKwFunction{GenerateTree}{GenerateSubtaskTree}
\SetKwFunction{GetModels}{GetModelsForSubtask}
\SetKwFunction{Backtrack}{BacktrackDependencies}

\textbf{Step 1: Generate Subtask Tree} \\
$G_{ss} \gets$ \GenerateTree{LLM, $x$, $u$, $\mathcal{S}$}\;

\textbf{Step 2: Build Tool Subgraph (TG)} \\
Initialize $G_{ts}$\;
\ForEach{subtask $s_i \in V_{ss}$}{
    $T_i \gets$ \GetModels{$MDT$, $s_i$}\;
    $G_{ti} \gets$ \Backtrack{$G_{td}$, $T_i$}\;
    Replace $s_i$ in $G_{ss}$ with $G_{ti}$ to construct $G_{ts}$\;
}

\Return{$G_{ts}$}\;
\end{algorithm}
% \vspace{-3em}

\section{A* Execution Strategy} \label{execution}

\begin{figure}[H]
    \centering
    \includegraphics[width=1\linewidth]{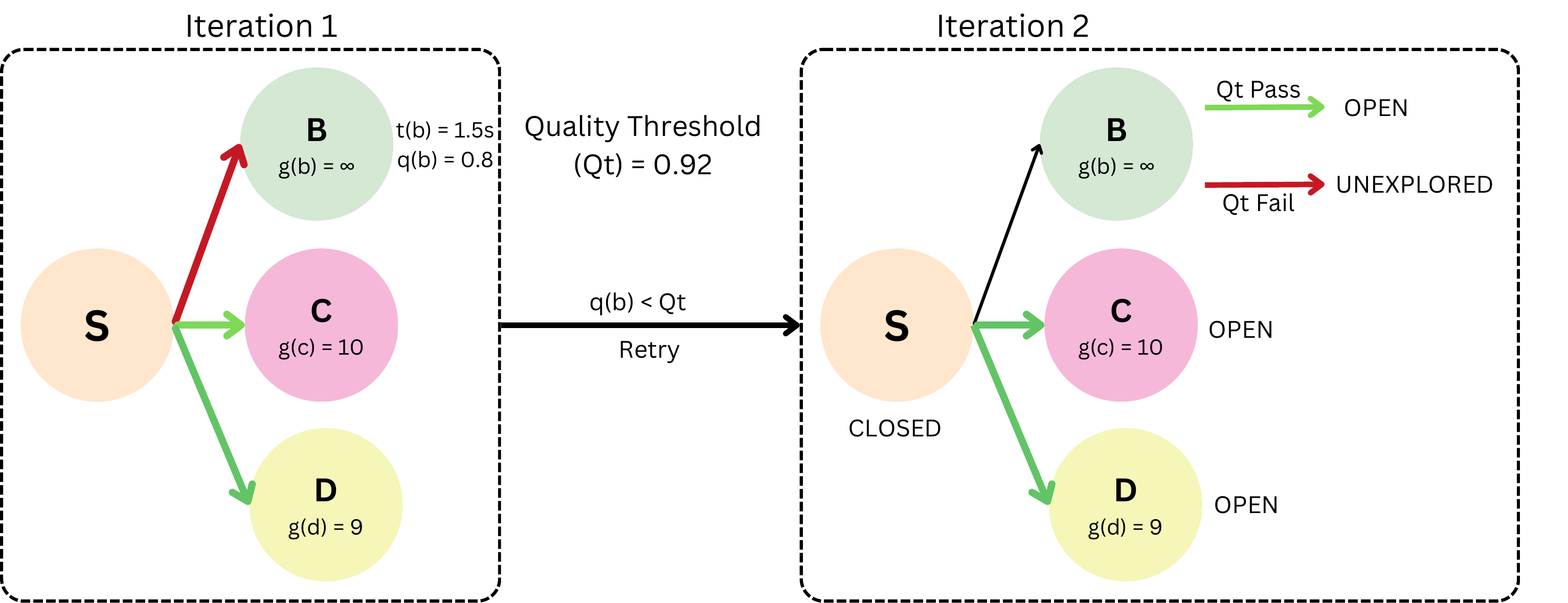}
    % \vspace{-1em}
    \caption{Visualization of the Retry Mechanism}
    \label{fig:enter-label}
    % \vspace{-1em}
\end{figure}

\ours{} initializes heuristic values using benchmark data and dynamically updates execution costs based on real-time performance. The A$^*$ search iteratively selects the node with the lowest \( f(x) \), explores its neighbors, and updates the corresponding values. If execution quality is below threshold, a retry mechanism adjusts parameters and re-evaluates \( g(x) \) (Figure \ref{fig:enter-label}). The process continues until a leaf node is reached.
By integrating precomputed heuristics with real-time cost updates, \ours{} efficiently balances execution time and quality. This adaptive approach ensures robust decision-making, outperforming existing agentic and non-agentic baselines in complex multimodal editing tasks.

\section{Correlation Analysis of CLIP Scores and Human Accuracy} \label{correlation}

\begin{figure}[H]
% \vspace{-0em}
    \centering
    \includegraphics[width=0.5\linewidth]{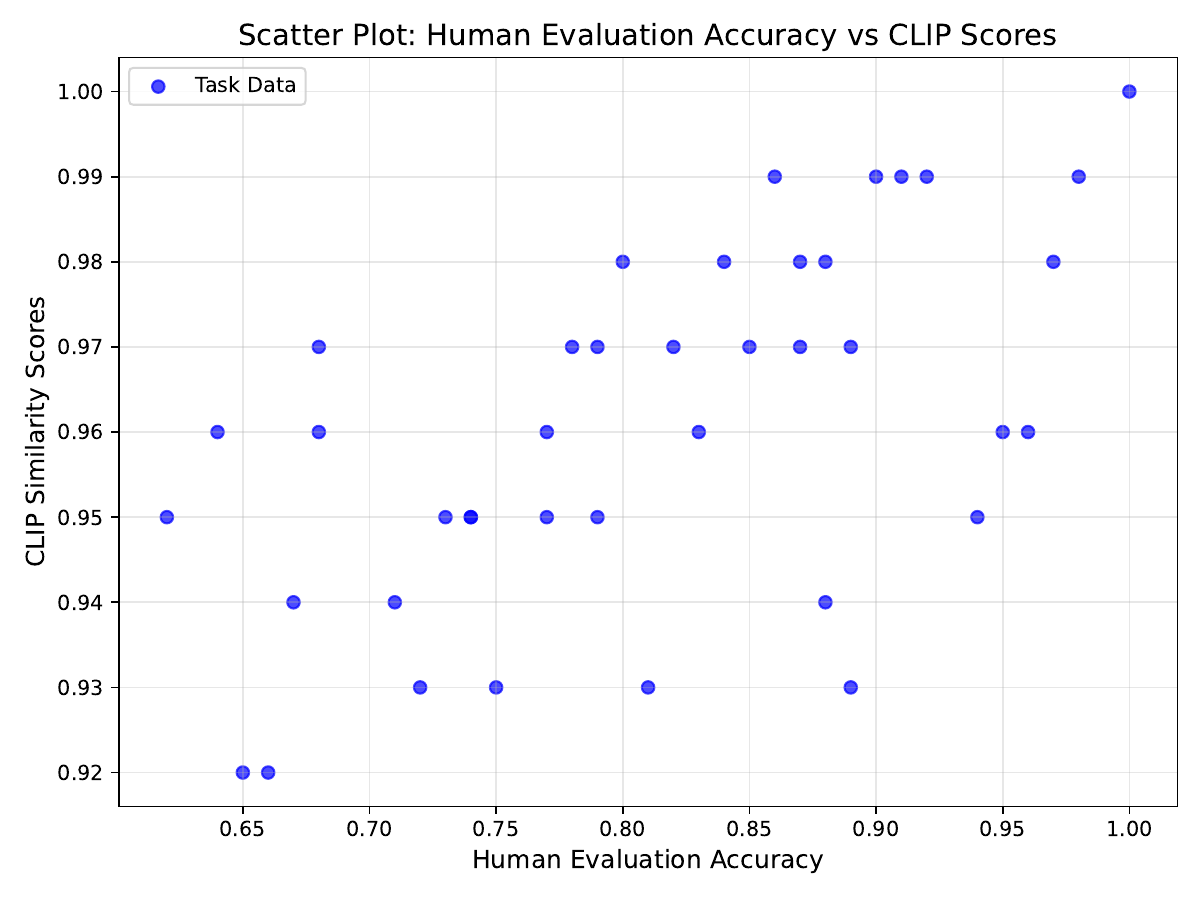}
    
    \caption{Scatter plot of CLIP scores vs. human accuracy across 40 tasks. The weak correlation (Spearman’s $\rho = 0.59$, Kendall’s $\tau = 0.47$) highlights CLIP’s limitations in capturing nuanced inaccuracies, particularly in complex, multi-step tasks.}
    \label{fig:scatter}
    % \vspace{-10em}
\end{figure}

We analyzed the correlation between CLIP similarity scores and human accuracy across 40 tasks to assess CLIP’s reliability in evaluating complex image-text edits. The scatter plot (Figure~\ref{fig:scatter}) illustrates the weak correlation, with Spearman’s $\rho = 0.59$ and Kendall’s $\tau = 0.47$, indicating that CLIP often fails to capture fine-grained inaccuracies. Despite assigning high similarity scores, CLIP struggles with detecting missing objects, distinguishing between multiple valid outputs, and recognizing context-dependent errors. Many instances where CLIP scored above 0.95 had human accuracy below 0.75, reinforcing the need for human evaluation in multimodal tasks. These findings highlight the limitations of CLIP as a standalone metric and emphasize the necessity of integrating human feedback for more reliable assessment.

% \vspace{-10em}
\section{LLM Prompt for Generating Subtask Tree} \label{prompt1}
\begin{tcolorbox}

\textbf{You are an advanced reasoning model responsible for decomposing a given image editing task into a structured subtask tree. Your task is to generate a well-formed subtask tree that logically organizes all necessary steps to fulfill the given user prompt. Below are key guidelines and expectations:}

\subsection{Understanding the Subtask Tree}

A subtask tree is a structured representation of how the given image editing task should be broken down into smaller, logically ordered subtasks. Each node in the tree represents an atomic operation that must be performed on the image. The tree ensures that all necessary operations are logically ordered, meaning a subtask that depends on another must appear after its dependency.

\subsection{Steps to Generate the Subtask Tree}

\begin{enumerate}
    \item \textbf{Step 1:} Identify all relevant subtasks needed to fulfill the given prompt.
    \item \textbf{Step 2:} Ensure that each subtask is logically ordered, meaning operations dependent on another should be placed later in the path.
    \item \textbf{Step 3:} Each subtask should be uniquely labeled based on the object it applies to and follow the format (Obj1 $\rightarrow$ Obj2) where Obj1 is replaced with Obj2. In case of recoloring, use (Obj $\rightarrow$ new color), while for removal, simply include (Obj) as the object being removed.
    \item \textbf{Step 4:} A tree may involve multiple correct paths where subtasks are independent of each other. In such cases, a subtask may appear twice in different parts of the tree. Number such occurrences distinctly, e.g., \texttt{Subtask1(1)}, \texttt{Subtask1(2)}, ensuring clarity.
    \item \textbf{Step 5:} Some tasks may have multiple valid approaches. For example, replacing a cat with a pink dog can be done in two ways:
          \begin{itemize}
              \item \texttt{Object Replacement (Cat $\rightarrow$ Pink Dog)}
              \item \texttt{Object Replacement (Cat $\rightarrow$ Dog)} $\rightarrow$ \texttt{Object Recoloration (Dog $\rightarrow$ Pink Dog)}
          \end{itemize}
\end{enumerate}

\subsection{Logical Constraints \& Dependencies}

\begin{itemize}
    \item Ensure proper ordering, e.g., if an object is replaced and then segmented, segmentation must follow replacement.
    \item Operations should be structured logically so that every subtask builds upon the previous one.
\end{itemize}

\subsection{Supported Subtasks}

Below is the complete list of available subtasks: Object Detection, Object Segmentation, Object Addition, Object Removal, Background Removal, Landmark Detection, Object Replacement, Image Upscaling, Image Captioning, Changing Scenery, Object Recoloration, Outpainting, Depth Estimation, Image Deblurring, Text Extraction, Text Replacement, Text Removal, Text Addition, Text Redaction, Question Answering Based on Text, Keyword Highlighting, Sentiment Analysis, Caption Consistency Check, Text Detection

\textbf{You must strictly use only these subtasks when constructing the tree.}

\subsection{Expected Output Format}

The model should output the subtask tree in structured JSON format, where each node contains:
\begin{itemize}
    \item \textbf{Subtask Name} (with object label if applicable)
    \item \textbf{Parent Node} (Parent subtask from which it depends)
    \item \textbf{Execution Order} (Logical flow of tasks)
\end{itemize}
\subsection{Example Inputs \& Expected Outputs}

\subsubsection{Example 1}
\end{tcolorbox}
\begin{tcolorbox}
\textbf{Input Prompt:} \textit{“Detect the pedestrians, remove the car and replacement the cat with rabbit and recolor the dog to pink.”}

\textbf{Expected Subtask Tree:}

\begin{verbatim}
"task": "Detect the pedestrians, remove the car and replacement the cat with 
\end{verbatim}
\begin{verbatim}
rabbit and recolor the dog to pink",
"subtask_tree": [
        {
            "subtask": "Object Detection (Pedestrian)(1)",
            "parent": []
        },
        {
            "subtask": "Object Removal (Car)(2)",
            "parent": ["Object Detection (Pedestrian)(1)"]
        },
        {
            "subtask": "Object Replacement (Cat -> Rabbit)(3)",
            "parent": ["Object Removal (Car)(2)"]
        },
        {
            "subtask": "Object Replacement (Cat -> Rabbit)(4)",
            "parent": ["Object Detection (Pedestrian)(1)"]
        },
        {
            "subtask": "Object Removal (Car)(5)",
            "parent": ["Object Replacement (Cat -> Rabbit)(4)"]
        },
        {
            "subtask": "Object Recoloration (Dog -> Pink Dog)(6)",
            "parent": ["Object Replacement (Cat -> Rabbit)(3)", \end{verbatim}
\begin{verbatim}                        "Object Removal (Car)(5)"]
        }
    ]
\end{verbatim}

\subsubsection{Example 2}

\textbf{Input Prompt:} \textit{“Update the closed signage to open while detecting the trash can and pedestrian crossing for better scene understanding. Also, remove the people for clarity.”}

\textbf{Expected Subtask Tree:}
\begin{verbatim}
"task": "Update the closed signage to open while detecting the trash can and \end{verbatim}
\begin{verbatim}pedestrian crossing for better scene understanding. Also, remove the people \end{verbatim}
\begin{verbatim}for clarity.",

"subtask_tree": [
        {
            "subtask": "Text Replacement (CLOSED -> OPEN)(1)",
            "parent": []
        },
        {
            "subtask": "Object Detection (Pedestrian Crossing)(2)",
            "parent": ["Text Replacement (CLOSED -> OPEN)(1)"]
        },
          \end{verbatim}
        \end{tcolorbox}
\begin{tcolorbox}
\begin{verbatim}
        {
            "subtask": "Object Detection (Trash Can)(3)",
            "parent": ["Text Replacement (CLOSED -> OPEN)(1)"]
        },
        {
            "subtask": "Object Detection (Pedestrian Crossing)(4)",
            "parent": ["Object Detection (Trash Can)(3)"]
        },
        {
            "subtask": "Object Detection (Trash Can)(5)",
            "parent": ["Object Detection (Pedestrian Crossing)(2)"]
        },
        {
            "subtask": "Object Removal (People)(6)",
            "parent": ["Object Detection (Pedestrian Crossing)(4)", \end{verbatim}
\begin{verbatim}                         "Object Detection (Trash Can)(5)"]
        }   
    ]

\end{verbatim}

\subsection{Final Task}

\textbf{Now, using the given input image and prompt, generate a well-structured subtask tree that adheres to the principles outlined above.}

\begin{itemize}
    \item Ensure logical ordering and clear dependencies.
    \item Label subtasks by object name where needed.
    \item Structure the output as a JSON-formatted subtask tree.
\end{itemize}

\textbf{Input Details:}
\begin{itemize}
    \item Image: \texttt{input\_image}
    \item Prompt: \texttt{User Prompt}
    \item Supported Subtasks: (See the list above)
\end{itemize}

\textbf{Now, generate the correct subtask tree. Before you generate the tree, ensure that for every possible path, all required subtasks are included and none are skipped.}

\end{tcolorbox}

\section{LLM Prompt for Getting Bounding Box and Text for Replacement}

\begin{tcolorbox}
\textbf{You are given an image containing text, where each word has associated bounding box coordinates. The existing text and their corresponding bounding boxes are as follows:}

\begin{itemize}
    \item \textbf{"THIS"}: (281,438,502,438,502,494,281,494)
    \item \textbf{"IS"}: (533,437,649,440,647,497,531,493)
    \item \textbf{"A"}: (667,444,734,444,734,492,667,492)
    \item \textbf{"NICE"}: (214,504,810,502,811,649,214,651)
    \item \textbf{"STREET"}: (68,674,915,640,924,859,77,893)
\end{itemize}

The user wants to replace this text with:

\begin{center}
    \textbf{"THIS IS NOT A NICE STREET"}
\end{center}

\subsection{Your Task}

You must determine which words in the image should be removed and which words need to be rewritten to ensure a smooth transition to the new text. The goal is to maintain spatial coherence while ensuring that the updated text fits naturally within the image.

\subsection{Guidelines for Text Replacement}

\begin{enumerate}
    \item \textbf{Identify Words to Remove:}
    \begin{itemize}
        \item Any word that needs to be replaced or modified should be marked for removal.
        \item If the new text introduces an additional word, the surrounding words should also be removed and rewritten to maintain proper spacing.
    \end{itemize}

    \item \textbf{Determine Placement for New Words:}
    \begin{itemize}
        \item If a word or phrase is being replaced (e.g., \texttt{"GOOD BOY"} → \texttt{"BAD GIRL"}), use a single bounding box that covers the area of both words instead of providing separate locations.
        \item If new words need to be inserted, ensure that adjacent words are also rewritten to provide sufficient space for readability.
        \item If the new text is longer than the original, adjust placements accordingly:
        \begin{itemize}
            \item Remove and rewrite words from the next or previous line if needed.
            \item If necessary, split the updated text into two separate lines and provide distinct bounding boxes for each.
        \end{itemize}
    \end{itemize}

    \item \textbf{Bounding Box Adjustments:}
    \begin{itemize}
        \item If text placement changes, the bounding box should be expanded or shifted to accommodate the new words.
        \item Ensure that all bounding boxes align with the natural flow of the text in the image.
    \end{itemize}
\end{enumerate}

\subsection{Example Case for Clarity}

\textbf{Input Scenario:}

    \textit{Original Text:} "I AM A GOOD BOY"
    \textit{Replacement Text:} "I AM A BAD GIRL"

\textbf{Expected Output:}
\begin{itemize}
    \item Remove: \texttt{"GOOD"} and \texttt{"BOY"}
    \vspace{-1em}
    \item Write: \texttt{"BAD GIRL"}
    \vspace{-1em}
    \item Bounding Box for "BAD GIRL": (Bounding box covering the area where "GOOD BOY" was originally written)
\end{itemize}

If \texttt{"BAD GIRL"} doesn’t fit naturally within the same space, adjust the bounding box or split it into multiple lines.
\end{tcolorbox}

\end{document}